\documentclass{article}

\PassOptionsToPackage{numbers,sort&compress}{natbib}
 \usepackage[preprint]{neurips_2026}
\usepackage[utf8]{inputenc}
\usepackage[T1]{fontenc}
\usepackage{hyperref}
\usepackage{url}
\usepackage{booktabs}
\usepackage{multirow}
\usepackage{amsfonts}
\usepackage{amsmath}
\usepackage{amssymb}
\usepackage{nicefrac}
\usepackage{microtype}
\usepackage[table]{xcolor}
\usepackage{graphicx}
\usepackage{float}
\usepackage{tikz}
\usepackage{cleveref}
\usetikzlibrary{arrows.meta}

\definecolor{BestMethodBlue}{RGB}{184,215,255}
\definecolor{SecondMethodBlue}{RGB}{229,241,255}
\newcommand{\bestcell}[1]{\cellcolor{BestMethodBlue}\textbf{#1}}
\newcommand{\secondcell}[1]{\cellcolor{SecondMethodBlue}#1}

\title{
PosteriorBench: From Point Estimates to Posterior Matching in Evaluating Generative Inverse Solvers \\
}

\newcommand*\samethanks[1][\value{footnote}]{\footnotemark[#1]}
\author{%
Jiachen Yao\textsuperscript{1}\thanks{Equal contribution.} \quad
Zi-Siang Hsu\textsuperscript{2}\samethanks \quad
Xi Deng\textsuperscript{1}\samethanks \quad
Aditi Gupta\textsuperscript{3}\\
\bfseries
Xin Ju\textsuperscript{4,5} \enspace
Sally M. Benson\textsuperscript{4,5} \enspace
Gege Wen\textsuperscript{6,5} \enspace
Anima Anandkumar\textsuperscript{1}\\[0.5ex]
\normalfont
\textsuperscript{1}California Institute of Technology \quad
\textsuperscript{2}National Taiwan University\\
\textsuperscript{3}Lawrence Berkeley National Laboratory\\
\textsuperscript{4}Department of Energy Science and Engineering, Stanford University\\
\textsuperscript{5}EarthFlow AI, Inc. \enspace
\textsuperscript{6}Department of Earth Sciences and Engineering, Imperial College London
}

\begin{document}

\maketitle

\begin{abstract}
Generative models are increasingly used to solve scientific inverse problems, but existing evaluations still focus primarily on whether a method can produce a single plausible reconstruction.
This is insufficient for ill-posed problems, where multiple solutions may be consistent with the same sparse or noisy observations.
In these settings, a method can achieve strong pointwise accuracy while still failing to capture the true posterior through mode collapse, overconfident uncertainty, or averaging incompatible solutions.
We introduce \textbf{PosteriorBench}, a benchmark for evaluating the \emph{distributional} accuracy of generative inverse solvers.
PosteriorBench evaluates four physics-based inverse problems: Darcy flow inversion, Poisson source recovery, carbon capture and storage, and light transport material inference.
For each task, we construct high-fidelity reference posteriors using computationally heavy but established procedures such as rejection sampling and Markov chain Monte Carlo, enabling direct assessment of whether solvers recover the full set of solutions rather than the single best sample.
We pair these references with a five-metric posterior evaluation suite: posterior-mean error, posterior-standard-deviation error, maximum mean discrepancy, sliced Wasserstein distance, and radially averaged power-spectrum error.
Together, these metrics assess pointwise accuracy, marginal uncertainty, distributional alignment, and global frequency fidelity.
The benchmark spans sparse sensing, low-resolution observations, nonlinear forward models, varying noise levels, and multimodal priors, with a unified pipeline for distribution matching and uncertainty quantification.
Our experiments reveal substantial distribution-matching gaps across current solvers, while showing that neural operators improve resolution robustness and that guidance weights and generation noise are key to posterior-variance calibration.
The code is available at \href{https://github.com/neuraloperator/PosteriorBench}{https://github.com/neuraloperator/PosteriorBench}.
\end{abstract}

\section{Introduction}
\label{sec:intro}

Inverse problems are central to scientific computing: one observes indirect, partial, or noisy measurements and seeks to infer the parameter field that produced them.
They arise in subsurface flow, optical imaging, fluid dynamics, and many other domains where direct measurement is expensive or impossible~\citep{tarantola2005inverse,mueller2012linear}.
The difficulty is not only that the forward physics may be nonlinear and expensive, but also that the inverse map is usually non-unique.
Bayesian inverse problems make this ambiguity explicit by placing a prior $p(x)$ over the unknown field $x$ and conditioning on the observed data $y_{\mathrm{obs}}$ through Bayes' rule,
\begin{equation}
    p(x \mid y_{\mathrm{obs}})
    \propto
    p(y_{\mathrm{obs}} \mid x)p(x).
    \label{eq:bayesian_inverse_posterior}
\end{equation}
Here the likelihood $p(y_{\mathrm{obs}} \mid x)$ is induced by the forward measurement model, while the prior $p(x)$ encodes data distribution assumptions~\citep{cotter2009bayesian,gelman1995bayesian}.

Recent generative models, especially diffusion and flow-based models, have made this Bayesian decomposition increasingly tractable.
Score-based generative models provide expressive priors for high-dimensional data~\citep{sohl2015deep,ho2020denoising,song2020score}, and diffusion-based sampling methods combine such priors with measurement likelihoods at inference time~\citep{chung2022diffusion,song2023pseudoinverse,wu2024principled,zhang2025improving,daras2024survey}.
In scientific settings, this area now intersects with neural operators and physics-informed learning~\citep{raissi2019physics,li2020fourier,kovachki2023neural,li2024physics}, leading to solvers that use joint coefficient-solution diffusion models, PDE residual guidance, or function-space formulations for physical fields~\citep{huang2024diffusionpde,shu2023physics,jacobsen2025cocogen,yao2025guided,wang2025fundiffdiffusionmodelsfunction,lin2026decoupled}.
These methods can produce ensembles, not merely point estimates, and are now being proposed as posterior samplers for PDE-constrained inverse problems.

Existing scientific inverse-problem benchmarks such as InverseBench~\citep{zheng2025inversebench} evaluate plug-and-play diffusion samplers across physical inverse problems, but still center on a single solution for each case.
More broadly, evaluation in this area is often anchored to single held-out ground-truth, which are insufficient for ill-posed problems.
A solver can match the observations and still underestimate posterior variance or blur out incompatible modes into an unphysical mean.

This gap motivates \textbf{PosteriorBench}, a benchmark for evaluating whether generative inverse solvers recover the \emph{posterior} they claim to sample from.
PosteriorBench focuses specifically on posterior matching: each benchmark case is paired with a transparent, computationally expensive reference posterior, so solvers can be evaluated as posterior samplers rather than only by their best reconstruction.
Generated ensembles are compared to the reference posteriors using metrics for distributional moments, spatial structure, observation consistency, and computational cost.
This framing is especially important for scientific settings, where posterior uncertainty informs decision making and downstream physical interpretation.

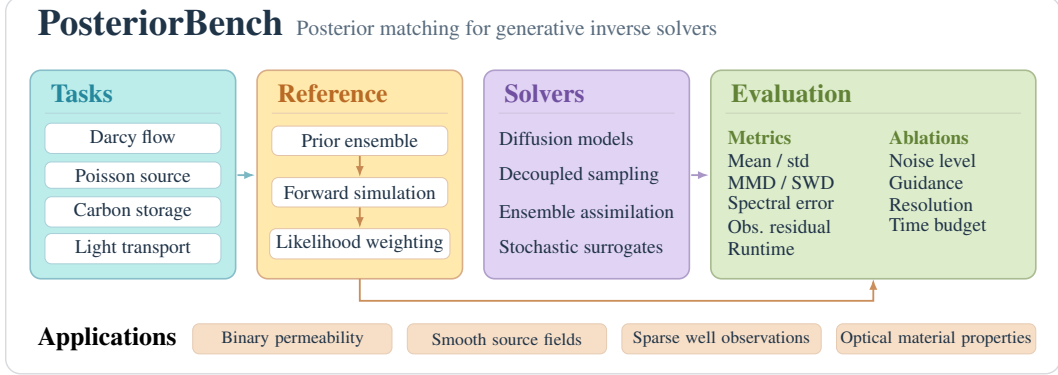
\begin{figure}[t]
  \centering
  \definecolor{pbInk}{HTML}{203448}
  \definecolor{pbMuted}{HTML}{586B7D}
  \definecolor{pbBlue}{HTML}{25889D}
  \definecolor{pbAmber}{HTML}{B96A25}
  \definecolor{pbPurple}{HTML}{7050A0}
  \definecolor{pbGreen}{HTML}{628335}
  \definecolor{pbCyanFill}{HTML}{BCEDEA}
  \definecolor{pbGoldFill}{HTML}{FFE9AC}
  \definecolor{pbPurpleFill}{HTML}{E1D3F3}
  \definecolor{pbGreenFill}{HTML}{DDECCB}
  \definecolor{pbPeachFill}{HTML}{F7DEC6}
  \resizebox{\textwidth}{!}{%
  \begin{tikzpicture}[
      font=\sffamily, text=pbInk,
      panel/.style={draw=#1!55, line width=0.75pt, rounded corners=5pt},
      heading/.style={font=\bfseries\large, anchor=west},
      item/.style={font=\small, anchor=west, inner sep=0pt},
      task/.style={draw=pbBlue!40, fill=white, rounded corners=3pt,
        minimum width=2.87cm, minimum height=0.43cm, font=\small, inner sep=3pt},
      refstep/.style={draw=pbAmber!40, fill=white, rounded corners=3pt,
        minimum width=2.87cm, minimum height=0.48cm, font=\small, inner sep=3pt},
      feature/.style={draw=pbAmber!35, fill=pbPeachFill, rounded corners=3pt,
        text width=3.15cm, align=center, minimum width=3.25cm,
        minimum height=0.49cm, font=\footnotesize, inner xsep=0.05cm, inner ysep=3pt},
      flowarrow/.style={-{Latex[length=1.8mm]}, line width=0.85pt, draw=pbMuted!75}
    ]
    \path[draw=pbInk!18, line width=0.65pt, rounded corners=7pt]
      (0,0) rectangle (17.2,6.15);
    \node[anchor=base west, font=\bfseries\LARGE] at (0.40,5.52) {PosteriorBench};
    \node[anchor=base west, font=\normalsize, text=pbMuted] at (4.62,5.52)
      {Posterior matching for generative inverse solvers};

    \path[panel=pbBlue, fill=pbCyanFill] (0.40,1.55) rectangle (3.75,4.95);
    \path[panel=pbAmber, fill=pbGoldFill] (4.10,1.55) rectangle (7.45,4.95);
    \path[panel=pbPurple, fill=pbPurpleFill] (7.80,1.55) rectangle (11.15,4.95);
    \path[panel=pbGreen, fill=pbGreenFill] (11.50,1.55) rectangle (16.80,4.95);
    \node[heading, text=pbBlue] at (0.61,4.58) {Tasks};
    \node[heading, text=pbAmber] at (4.31,4.58) {Reference};
    \node[heading, text=pbPurple] at (8.01,4.58) {Solvers};
    \node[heading, text=pbGreen] at (11.71,4.58) {Evaluation};
    \draw[pbBlue!35] (0.64,4.25) -- (3.51,4.25);
    \draw[pbAmber!35] (4.34,4.25) -- (7.21,4.25);
    \draw[pbPurple!35] (8.04,4.25) -- (10.91,4.25);
    \draw[pbGreen!35] (11.74,4.25) -- (16.56,4.25);

    \node[task] at (2.075,3.84) {Darcy flow};
    \node[task] at (2.075,3.24) {Poisson source};
    \node[task] at (2.075,2.64) {Carbon storage};
    \node[task] at (2.075,2.04) {Light transport};

    \node[refstep] (prior) at (5.775,3.80) {Prior ensemble};
    \node[refstep] (forward) at (5.775,2.96) {Forward simulation};
    \node[refstep] (weight) at (5.775,2.12) {Likelihood weighting};
    \draw[flowarrow, draw=pbAmber!80] (prior.south) -- (forward.north);
    \draw[flowarrow, draw=pbAmber!80] (forward.south) -- (weight.north);

    \node[item] at (8.04,3.84) {Diffusion models};
    \node[item] at (8.04,3.24) {Decoupled sampling};
    \node[item] at (8.04,2.64) {Ensemble assimilation};
    \node[item] at (8.04,2.04) {Stochastic surrogates};

    \node[item, font=\bfseries\small, text=pbGreen] at (11.77,3.86) {Metrics};
    \node[item, font=\bfseries\small, text=pbGreen] at (14.40,3.86) {Ablations};
    \foreach \y/\label in {3.48/{Mean / std},3.12/{MMD / SWD},2.76/{Spectral error},2.40/{Obs. residual},2.04/{Runtime}} {
      \node[item] at (11.77,\y) {\label};
    }
    \foreach \y/\label in {3.48/{Noise level},3.12/{Guidance},2.76/{Resolution},2.40/{Time budget}} {
      \node[item] at (14.40,\y) {\label};
    }
    \draw[flowarrow, draw=pbBlue!60] (3.78,3.24) -- (4.07,3.24);
    \draw[flowarrow, draw=pbPurple!60] (11.18,3.24) -- (11.47,3.24);
    \draw[flowarrow, draw=pbAmber!75] (5.775,1.55) -- (5.775,1.19)
      -- (14.15,1.19) -- (14.15,1.55);

    \node[anchor=west, font=\bfseries\large, text=black] at (0.4,0.57) {Applications};
    \node[feature] at (4.675,0.57) {Binary permeability};
    \node[feature] at (8.175,0.57) {Smooth source fields};
    \node[feature] at (11.675,0.57) {Sparse well observations};
    \node[feature] at (15.175,0.57) {Optical material properties};
  \end{tikzpicture}%
  }
  \caption{Overview of PosteriorBench.
  Across four scientific inverse tasks, each case pairs a fixed observation with a weighted reference posterior and a solver-generated ensemble.
  Evaluation compares posterior moments, distributional alignment, spatial structure, observation consistency, and computational cost.
  }
  \label{fig:teaser}
  \vspace{-1em}
\end{figure}

The benchmark contains four tasks from different scientific domains: Darcy flow inversion, Poisson source recovery, carbon capture and storage (CCS), and light transport material inference (LTMI).
Together they cover binary and smooth priors, sparse point and column observations, nonlinear physical maps, low-resolution measurements, and structured scientific priors.
The CCS task is a high-impact subsurface monitoring setting, where permeability fields must be inferred from sparse well observations of CO$_2$ dynamics; reducing the number of wells can substantially lower field intervention and monitoring cost.
Ensemble methods such as ES-MDA remain important domain baselines for this inverse problem~\citep{Emerick2013b,Jung2018}.
The LTMI task adds a two-layer radiative inverse problem motivated by atmospheric imaging, optical tomography, and nondestructive material inspection, where optical measurements must be translated into plausible internal material structure.

From the benchmark, we find that function-space diffusion samplers such as DDIS~\citep{lin2026decoupled} and FunDPS~\citep{yao2025guided} are strong posterior samplers across several scientific inverse tasks.
We also find that posterior-mean error alone can be misleading: a solver may place the ensemble center near the reference mean while still misrepresenting posterior spread.
This makes moment consistency a joint requirement, where posterior mean and standard deviation errors should be interpreted together and, when possible, alongside distributional metrics such as MMD and SWD.

We further compare posterior metrics against a traditional pointwise reconstruction metric.
This comparison reveals that aggressively fitting a single reference field can degrade posterior structure: low pointwise error can coincide with poor posterior variance, distributional mismatch, or distorted spatial statistics.
Through diagnostics based on the governing PDE constraint and the latent GRF smoothness parameter, we find that posterior metrics better reflect physically meaningful recovery than pointwise error alone.

We also study how solvers respond to different observation noise levels.
For a Gaussian observation model, guidance weights should, in theory, scale with the inverse observation-noise variance.
Our sweeps show that learned samplers do not resolve posterior calibration by this scaling alone: stronger guidance improves observation consistency but often underestimates posterior variance, while weaker guidance preserves diversity at the cost of a biased or weakly conditioned posterior mean.
This mean--variance tradeoff motivates conditioning mechanisms that can calibrate posterior mean and uncertainty jointly, rather than relying on a single weight.
\Cref{sec:exp} details more ablation studies.

Our contributions are threefold.
First, we introduce a distribution-centered evaluation protocol for generative scientific inverse solvers that explicitly incorporates reference posterior construction and validation.
Second, we organize four diverse inverse tasks under a common distributional evaluation suite, and we release code for reproducing these experiments and evaluating new solvers on PosteriorBench.
Third, our experiments and ablations identify persistent distribution-matching gaps and insights into future method development in physics-based inverse problems.

\section{Preliminaries}
\label{sec:prelim}

An inverse problem seeks an unknown physical field $x \in \mathcal{X}$ from indirect observations $y_{\mathrm{obs}} \in \mathcal{Y}$ produced by a forward map $\mathcal{A}:\mathcal{X}\to\mathcal{Y}$, such as a PDE solver, reservoir simulator, or radiative transport model.
Because observations are often sparse or noisy and $\mathcal{A}$ is often many-to-one, the scientifically relevant target is usually not a single reconstruction but the posterior distribution (\ref{eq:bayesian_inverse_posterior}).

In PosteriorBench, a \emph{task} or \emph{problem} denotes a family of inverse problems, while a \emph{case} denotes one observation-conditioned posterior recovery setting.
For each case, a solver receives $y_{\mathrm{obs}}$ and returns \emph{samples} intended to approximate \Cref{eq:bayesian_inverse_posterior}.

Diffusion models learn the prior by perturbing clean samples $x_0\sim p(x)$ into noisy variables $x_t$ and training a network $s_{\theta}(x_t,t)$ to approximate the score $\nabla_{x_t}\log p_t(x_t)$~\citep{sohl2015deep,ho2020denoising,song2020score}.
For a forward noising SDE $\mathrm{d}x_t=f(x_t,t)\mathrm{d}t+g(t)\mathrm{d}w_t$, reverse sampling follows
\begin{equation}
    \mathrm{d}x_t
    =
    \left[
        f(x_t,t)-g(t)^2 s_{\theta}(x_t,t)
    \right]\mathrm{d}t
    +
    g(t)\mathrm{d}\bar{w}_t,
    \qquad
    t:T\to0,
    \label{eq:reverse_diffusion_sde}
\end{equation}
where $\bar{w}_t$ denotes Brownian motion in reverse time.
To condition such a prior on observations, diffusion posterior sampling~\cite{chung2022diffusion} uses the posterior-score decomposition
\begin{equation}
    \nabla_x \log p(x\mid y_{\mathrm{obs}})
    =
    \nabla_x \log p(x)
    +
    \nabla_x \log p(y_{\mathrm{obs}}\mid x),
    \label{eq:posterior_score_decomposition}
\end{equation}
combining the learned diffusion score with an inference-time likelihood or guidance term~\citep{chung2022diffusion,song2023pseudoinverse,wu2024principled,zhang2025improving,daras2024survey}.
For the additive Gaussian observation model
\begin{equation}
    y_{\mathrm{obs}} = \mathcal{A}(x) + \eta,
    \qquad
    \eta \sim \mathcal{N}(0,\sigma_y^2 I),
    \label{eq:gaussian_observation_model}
\end{equation}
the likelihood term is
\begin{equation}
    \nabla_x \log p(y_{\mathrm{obs}}\mid x)
    =
    -\frac{1}{2\sigma_y^2}
    \nabla_x
    \left\|
        \mathcal{A}(x)-y_{\mathrm{obs}}
    \right\|_2^2.
    \label{eq:gaussian_likelihood_score}
\end{equation}
Practical diffusion posterior samplers usually apply this gradient to a denoised estimate $\hat{x}_0(x_t)$ during the reverse diffusion trajectory, so that each reverse step balances prior against agreement with the observed data.
In scientific inverse problems, $\mathcal{A}$ may be differentiable, replaced by a neural-operator surrogate or a physics residual, leading to variants such as joint coefficient-solution diffusion, function-space guidance, and decoupled prior sampling~\citep{huang2024diffusionpde,yao2025guided,wang2025fundiffdiffusionmodelsfunction,lin2026decoupled}.

\section{PosteriorBench}
\label{sec:dataset}

\begin{figure}[ht]
  \centering
  \includegraphics[width=1.0\textwidth]{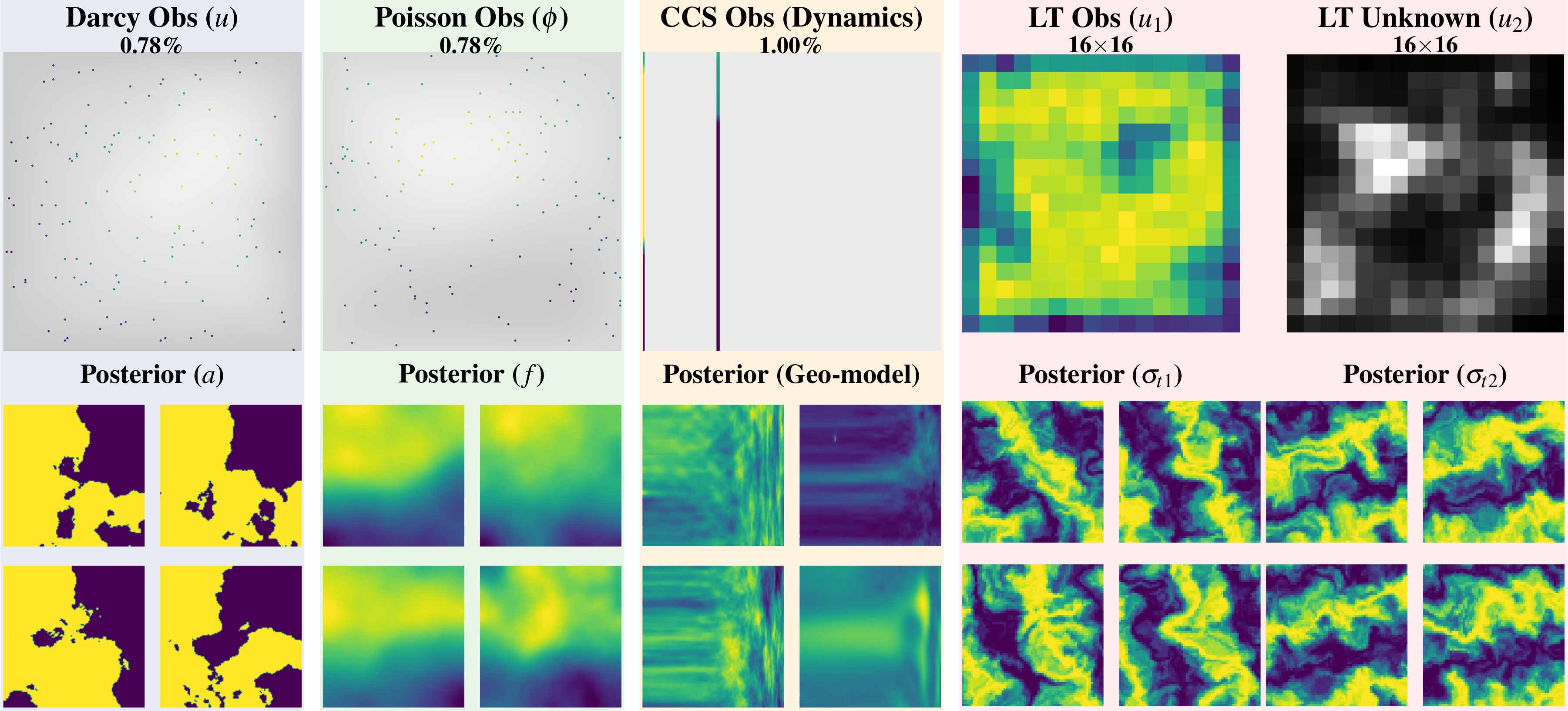}
  \caption{Representative cases from PosteriorBench.
  The top row shows the observation available to the inverse solver: sparse pressure measurements for Darcy flow and Poisson source recovery, sparse well-column measurements for CCS, and a low-resolution filter for light-transport material inference (LTMI).
  The bottom row shows multiple samples from the corresponding reference posterior.
  }
  \label{fig:representative_cases}
\end{figure}

PosteriorBench is designed to evaluate whether a generative inverse solver recovers the \emph{posterior distribution} rather than only a single accurate reconstruction, as illustrated by the representative cases in \Cref{fig:representative_cases}.
The benchmark emphasizes settings where ambiguity is intrinsic, posterior mass is multi-mode, and uncertainty quantification is scientifically meaningful.
At a high level, the benchmark is built around three principles: controllable inverse problems with known physics, high-fidelity reference posteriors, and evaluation metrics that compare distributions rather than point estimates.

\subsection{Benchmark Task Design}
\label{subsec:datasets}

PosteriorBench is organized around inverse problems in which sparse or aggregated measurements admit multiple physically plausible latent fields.
We choose four tasks that vary along the main axes that affect posterior recovery: the prior over unknown fields, the observation pattern, the forward physics, and the dominant source of posterior ambiguity.
\Cref{tab:task_features} summarizes these design choices, while the following subsections describe the construction of each task.

\begin{table}[h]
  \centering
  \caption{Keyword feature comparison of the PosteriorBench tasks. Each row summarizes the prior, observation pattern, forward model, and dominant ambiguity source used in the benchmark.}
  \label{tab:task_features}
  \footnotesize
  \setlength{\tabcolsep}{2pt}
  \begin{tabular*}{\textwidth}{@{\extracolsep{\fill}}p{0.25\textwidth}p{0.18\textwidth}p{0.15\textwidth}p{0.15\textwidth}p{0.15\textwidth}}
    \toprule
    Task & Prior type & Observation & Forward map & Ambiguity \\
    \midrule
    Darcy flow inversion & Binary & Sparse sensors & PDE solver & Phase layout \\
    Poisson source recovery & Smooth GRF & Sparse sensors & PDE solver & Spectral modes \\
    CO$_2$ capture and storage & Geostat. multimodal & Well columns & Reservoir sim. & Plume uncertainty \\
    Light transp. material inference & Cloud texture & Low-res optical & Radiative & Material layout \\
    \bottomrule
  \end{tabular*}
\end{table}

\subsubsection{Darcy Flow Inversion}
Darcy flow is a canonical elliptic inverse problem for porous-media modeling~\citep{yao2025guided}.
On $\Omega=(0,1)^2$, we consider the steady equation
\begin{equation}
    -\nabla \cdot (a(x)\nabla u(x)) = 1,\quad x\in\Omega,
    \qquad
    u|_{\partial\Omega}=0,
\end{equation}
where $a(x)$ is the unknown permeability or conductivity field and $u(x)$ is the corresponding pressure response.
This setup uses constant unit forcing and homogeneous Dirichlet boundary conditions, matching the standard Darcy data-generation convention.
The inverse task is to recover the posterior distribution of $a$ from sparse point observations of $u$.

Following the standard Darcy construction used in operator-learning benchmarks~\citep{li2020fourier}, we sample a Gaussian random field (GRF) and threshold it into binary high- and low-conductivity phases, with representative values $12,3$.
This prior creates discontinuous material interfaces and channel-like structures.
Sparse observations constrain the induced flow field, but they do not uniquely determine the underlying phase layout: multiple connected high-conductivity pathways can produce similar pressure measurements at the observed locations.
The Darcy task therefore evaluates whether a solver captures posterior uncertainty rather than a single plausible reconstruction.

\subsubsection{Poisson Source Recovery}
To assess solvers in a continuous and more spectrally variable setting, we design the Poisson source recovery task, governed by the equation 
\vspace{-0.5em}
\begin{equation}
    -\nabla^2 \phi = f,
\end{equation}
where $f$ is the source term and $\phi$ is the potential.
The goal is to recover the source term $f$ from sparse observations of the potential $\phi$.
The workflow for posterior construction follows the Darcy task using sparse observation, but the prior introduces broader variation across cases.

The source terms $f$ are drawn from Gaussian random fields with varying correlation length $\tau$ and smoothness $\alpha$.
Sampling these hyperparameters independently for each case increases the structural diversity and spectral variability of the fields.
The Poisson task therefore evaluates whether inverse solvers generalize across a broad family of prior spectra rather than overfitting to a single fixed prior.

\subsubsection{Carbon Capture and Storage}
Carbon capture and storage (CCS) requires reliable characterization of subsurface heterogeneity in order to forecast and monitor CO$_2$ plume migration, pressure buildup, and storage security~\citep{pacala2004stabilization}.
In PosteriorBench, the CCS task is formulated as a Bayesian inverse problem over a static permeability field.
Let $m$ denote the subsurface geomodel and let $s = F(m)$ denote the dynamic CO$_2$ saturation response after injection.
Given sparse observations collected at monitoring wells, the goal is to recover the posterior distribution $p(m \mid y_{\mathrm{obs}})$ rather than a single calibrated permeability map.

The task follows the sparse-monitoring regime used in recent function-space diffusion work for CCS~\citep{ju2026funddps}.
Permeability realizations are generated from geostatistical priors using SGeMS-style simulation~\citep{sgems}, and the corresponding CO$_2$ saturation fields are produced with a high-fidelity reservoir simulator such as ECLIPSE~\citep{eclipse}.
Observations are represented as vertical strip or column patterns that mimic well measurements, so the inverse solver observes only a small fraction of the spatial domain.
This setting is deliberately challenging: many geomodels can match the same well measurements, and the posterior can contain substantial spatial uncertainty away from the wells.

For this task, the reference posterior is constructed by drawing a large candidate pool from the geological prior and accepting or weighting candidates according to their mismatch to the observed saturation response under the forward model or a validated neural-operator surrogate.
The benchmark reports posterior matching against this reference ensemble, while also tracking observation consistency and reference-posterior validation diagnostics.
Further details on the dataset, simulation, and posterior construction are provided in \Cref{app:ccs_setup}.
We note that ES-MDA is a widely used data-assimilation method in reservoir characterization, thus we include it in our baselines~\citep{Emerick2013b,Jung2018}.

\subsubsection{Light transport material inference (LTMI)}
Inverse light transport through multilayer scattering materials arises in a range of applications, including atmospheric retrieval of cloud and aerosol structure, diffuse optical tomography of layered biological tissue, and nondestructive optical inspection of semitransparent materials.
In these settings, the goal is to reconstruct spatially varying optical properties, such as extinction or scattering coefficients, from sparse observations.
This inverse problem is challenging because the unknown field is high-dimensional, while the available measurements are often severely limited in viewpoint or acquisition time due to hardware and practical constraints.
For example, in atmospheric imaging, it is often difficult to obtain simultaneous multi-view observations of the same cloud field.

In our dataset, each sample consists of a two-layer participating medium arranged along the viewing direction, where each layer contains a spatially varying density field with either cloud-like or cellular-like morphology.
We then consider two kinds of measurement, reflectance fields and transmittance fields, each with low-resolution or pointwise observation.
The goal is to infer the extinction coefficients of the participating material, which is ambiguous because different arrangements of scattering material can produce similar aggregate measurements under limited observation.

\subsection{Reference Posterior Construction}

A central feature of PosteriorBench is that each benchmark case is paired with a high-fidelity reference posterior.
Depending on the problem structure, this reference distribution is obtained through slow but reliable procedures such as rejection sampling and Markov chain Monte Carlo.
For rejection-sampling-based tasks, we draw a large candidate pool from the prior and simulate the observation process for each candidate.
We specify the assumed Gaussian observation noise level $\sigma$ and use it to assign soft likelihood weights
\begin{equation}
    w_i \propto \exp\left(
        -\frac{\|\mathcal{A}(x_i)-y_{\mathrm{obs}}\|_2^2}{2\sigma^2}
    \right).
\end{equation}
For efficient metric computation, we set threshold $\epsilon = 3\sigma$ and randomly keep 100 samples whose observation mismatch is below the $\epsilon$-threshold.
We normalize the weights over the retained ensemble so that $\sum_i w_i = 1$.
Before using these reference posteriors as evaluation targets, we validate that they are stable and observation-consistent.
The validation protocol is provided in \Cref{app:reference_strength_validation}.

\subsection{Evaluation Metrics}
\label{subsec:metrics}

To evaluate inverse solvers as posterior samplers, we use a metric suite that compares generated ensembles with weighted reference posteriors.
The reference distribution is represented as a weighted empirical distribution $\mathcal{P}_{ref} = \{(x_i, w_i)\}_{i=1}^M$, where weights $w$ are derived from rejection sampling or importance sampling.
In contrast, the solvers under evaluation typically produce an unweighted ensemble of samples $\mathcal{P}_{gen} = \{x'_j\}_{j=1}^N$. 
Considering the asymmetry, our metrics are as follows.

\subsubsection{Marginal Moment Consistency}

We first check whether the generated ensemble matches the low-order posterior statistics often used in downstream scientific analysis.
The posterior mean measures the expected reconstructed field, while the marginal standard deviation measures the magnitude of pointwise uncertainty.
For generated samples $\mathcal{P}_{gen}=\{x'_j\}_{j=1}^N$ and a weighted reference posterior $\mathcal{P}_{ref}=\{(x_i,w_i)\}_{i=1}^M$, we define
\begin{equation}
    \bar{\mu}_{gen} = \frac{1}{N}\sum_{j=1}^N x'_j,
    \qquad
    \bar{\mu}_{ref} = \sum_{i=1}^M w_i x_i,
\end{equation}
and compute the relative mean error and, similarly, the std error:
\begin{equation}
    \mathrm{Err}_{\mu}
    =
    \frac{\|\bar{\mu}_{gen} - \bar{\mu}_{ref}\|_2}
         {\|\bar{\mu}_{ref}\|_2},
    \qquad
    \mathrm{Err}_{\sigma}
    =
    \frac{\|\bar{\sigma}_{gen} - \bar{\sigma}_{ref}\|_2}
         {\|\bar{\sigma}_{ref}\|_2}.
\end{equation}
We emphasize that the relative mean error differs from the mean squared error computed against a single target.
These moment errors provide a straightforward check on whether a solver recovers not only the expected field, but also where the inverse problem remains uncertain.

\subsubsection{Distributional and Geometric Alignment}
To evaluate alignment beyond marginal moments, we use two complementary distributional measures.
The first compares samples directly in the physical field space through a characteristic kernel, while the second compares one-dimensional projections after field-scale normalization.

\paragraph{Maximum Mean Discrepancy (MMD)} 
We use MMD to detect discrepancies in higher-order spatial statistics.
Let the generated ensemble have normalized weights $w'_j=1/N$ and let the reference posterior have normalized weights $w_i$.
The squared MMD is
\begin{equation}
\mathrm{MMD}^2
=
\sum_{i,i'} w_i w_{i'} k(x_i,x_{i'})
+
\sum_{j,j'} w'_j w'_{j'} k(x'_j,x'_{j'})
-
2\sum_{i,j} w_i w'_j k(x_i,x'_j).
\end{equation}
The reported MMD is the square root of this quantity.
The kernel is a multi-scale RBF kernel
\begin{equation}
    k(x,y)
    =
    \frac{1}{|\mathcal{S}|}
    \sum_{s\in\mathcal{S}}
    \exp\left(
        -\frac{\|x-y\|_2^2}{ds}
    \right),
    \qquad
    \mathcal{S}=\{0.2,0.5,1.0,2.0,5.0\},
\end{equation}
where $d$ is the median squared distance between generated and reference samples. 
This multi-scale kernel makes the statistic sensitive to both coarse structural shifts and finer spatial discrepancies.

\paragraph{Sliced Wasserstein Distance (SWD)}
To evaluate geometric proximity, we compute a sliced Wasserstein distance between generated and reference ensembles.
Because the tasks have different physical units and dynamic ranges, SWD is computed after field-wise z-score normalization.
We then project the normalized fields onto smooth random directions rather than i.i.d. pixel-wise Gaussian vectors to avoid local cancellation.
This GRF projection distribution favors spatially coherent test functions, making SWD aligned with physical field discrepancies.
For each direction $v_\ell$, we compute the weighted one-dimensional Wasserstein distance between projected samples:
\begin{equation}
    \mathrm{SWD}
    =
    \frac{1}{L}
    \sum_{\ell=1}^{L}
    W_1\left(
        \sum_{j=1}^{N} w'_j \delta_{\langle \tilde{x}'_j,v_\ell\rangle},
        \sum_{i=1}^{M} w_i \delta_{\langle \tilde{x}_i,v_\ell\rangle}
    \right).
\end{equation}
We use $L=128$ projections with default GRF parameters $\alpha=2$ and $\tau=3$.

\subsubsection{Spectral Analysis}
In physical inverse problems, the statistical texture and energy distribution across scales are of informative as well. We therefore compute the \textbf{Radially Averaged Power Spectrum (RAPS)} to evaluate spectral consistency. 

For each sample $x \in \mathbb{R}^{H \times W}$, we first compute its 2D power spectrum $P_x(q):=|\mathcal{F}(x)(q)|^2$ using the Discrete Fourier Transform (DFT).
The 2D spectrum is then mapped to a 1D representation $S(k)$ by averaging the power density within radial bins $k = \sqrt{k_x^2 + k_y^2}$.
For the reference distribution $\mathcal{P}_{ref}$, the ensemble spectrum is defined as the weighted average:
\begin{equation}
    S_{ref}(k)
    =
    \sum_{i=1}^M w_i R_{x_i}(k),
    \qquad
    R_{x_i}(k)
    =
    \frac{1}{|B_k|}
    \sum_{q\in B_k} P_{x_i}(q),
\end{equation}
For generated samples, we define $S_{gen}(k)=\frac{1}{N}\sum_{j=1}^N R_{x'_j}(k)$ and report the geometric mean of per-bin relative errors
\begin{equation}
    \mathrm{Err}_{\mathrm{spec}}
    =
    \exp\left(
        \frac{1}{|\mathcal{K}|}
        \sum_{k\in\mathcal{K}}
        \log\left(
            \frac{|S_{gen}(k)-S_{ref}(k)|}{|S_{ref}(k)|}
        \right)
    \right).
\end{equation}
Here $\mathcal{K}$ denotes the valid frequency bins.
The geometric mean is used to avoid overemphasizing high-frequency bins with low power.
This metric assesses whether the solver preserves the physical energy cascade and avoids common pitfalls such as over-smoothing, which is often hard to tell from spatial-domain metrics.

\section{Experiments}
\label{sec:exp}

\providecommand{\emptyresult}{\phantom{0.000}}

\subsection{Methods}

We evaluate eight probabilistic inverse solvers on all four PosteriorBench tasks: ECI-sampling~\citep{cheng2025gradientfree}, DiffusionPDE~\citep{huang2024diffusionpde}, FunDPS~\citep{yao2025guided}, Fun-DDPS~\citep{ju2026funddps}, DDIS~\citep{lin2026decoupled}, FunDiff~\citep{wang2025fundiffdiffusionmodelsfunction}, ES-MDA~\citep{Emerick2013b,Jung2018}, and FNO with MC Dropout.
Fun-DDPS and DDIS instantiate decoupled posterior sampling strategies that pair a learned prior with surrogate- or physics-based likelihood guidance, while ES-MDA provides a classical ensemble data-assimilation baseline and FNO with MC Dropout provides a direct neural uncertainty baseline.

\subsection{Main Results}

\Cref{tab:main_results} summarizes the quantitative evaluation of posterior-generating solvers across the benchmark tasks.
\Cref{app:additional_results} provides full guidance-weight sweeps and case-level diagnostics, pairwise metric analyses, and standard deviations across cases.

\begin{table}[t]
  \centering
  \caption{Main benchmark results across all PosteriorBench tasks.
  Lower is better for all metrics.
  Best and second-best results within each task and posterior-quality metric are highlighted in dark and light blue, respectively; runtime is rounded up to 0.1 min.
  }
  \label{tab:main_results}
  \resizebox{\textwidth}{!}{%
  \begin{tabular}{llcccccc}
    \toprule
    Task & Method & Mean Error $\downarrow$ & Std Error $\downarrow$ & MMD $\downarrow$ & SWD $\downarrow$ & Spectral Error $\downarrow$ & Time $\downarrow$ \\
    \midrule
    \multirow{8}{*}{Darcy} & ECI & 0.5112 & 1.4975 & 0.6457 & 55.2020 & 0.2007 & 3.6 min \\
    & ES-MDA & 0.4519 & 1.7547 & 0.5812 & 48.4836 & 0.3481 & \textbf{0.1 min} \\
    & FunDPS & 0.0804 & 0.4354 & \secondcell{0.2040} & 3.0778 & 0.2016 & 8.8 min \\
    & Fun-DDPS & \secondcell{0.0767} & \secondcell{0.4178} & 0.2131 & \bestcell{2.7840} & \secondcell{0.1222} & 7.1 min \\
    & DiffusionPDE & 0.1009 & 0.7871 & 0.2268 & 6.7372 & 0.3252 & 40.3 min \\
    & DDIS & \bestcell{0.0620} & \bestcell{0.4130} & \bestcell{0.1997} & \secondcell{2.9236} & 0.1676 & 14.2 min \\
    & FunDiff & 0.1203 & 0.5900 & 0.3229 & 4.0833 & 0.1225 & 0.5 min \\
    & MC-dropout & 0.1990 & 0.9509 & 0.6728 & 6.0297 & \bestcell{0.1123} & \textbf{0.1 min} \\
    \midrule
    \multirow{8}{*}{Poisson} & ECI & 3.4434 & 8.7178 & 0.7623 & 69.5794 & 35.0923 & 0.8 min \\
    & ES-MDA & 1.0933 & 5.6689 & 0.6370 & 41.5815 & 7.7886 & \textbf{0.1 min} \\
    & FunDPS & 0.4720 & 0.5862 & 0.6692 & 4.3900 & 0.2303 & 9.4 min \\
    & Fun-DDPS & \secondcell{0.1839} & \secondcell{0.4329} & \secondcell{0.3576} & \secondcell{2.4304} & \bestcell{0.0752} & 7.3 min \\
    & DiffusionPDE & 0.3319 & 0.7004 & 0.6814 & 4.2661 & 0.1907 & 40.0 min \\
    & DDIS & \bestcell{0.1413} & \bestcell{0.3073} & \bestcell{0.2570} & \bestcell{1.8568} & \secondcell{0.1760} & 15.3 min \\
    & FunDiff & 1.0529 & 4.3679 & 0.6363 & 34.8920 & 4.0411 & 0.5 min \\
    & MC-dropout & 0.3495 & 0.8837 & 0.7157 & 4.4383 & 0.2053 & \textbf{0.1 min} \\
    \midrule
    \multirow{8}{*}{CCS} & ECI & 0.6753 & 1.5330 & 0.4991 & 10.5775 & 88.8274 & 0.3 min \\
    & ES-MDA & 0.2457 & 1.0903 & 0.2941 & 8.0556 & 14.5449 & \textbf{0.1 min} \\
    & FunDPS & \bestcell{0.1419} & \bestcell{0.2952} & \bestcell{0.2018} & \bestcell{4.2564} & \bestcell{0.6370} & 6.3 min \\
    & Fun-DDPS & 0.4659 & 1.0305 & \secondcell{0.2410} & 6.9947 & 14.7840 & 6.2 min \\
    & DiffusionPDE & 0.2015 & \secondcell{0.3549} & 0.2781 & \secondcell{5.8572} & 0.7294 & 34.0 min \\
    & DDIS & 0.3277 & 0.4332 & 0.3007 & 7.3551 & 2.2043 & 9.2 min \\
    & FunDiff & \secondcell{0.1792} & 0.5195 & 0.3647 & 7.1329 & 0.6896 & 1.5 min \\
    & MC-dropout & 0.2574 & 0.8926 & 0.6844 & 13.5715 & \secondcell{0.6555} & \textbf{0.1 min} \\
    \midrule
    \multirow{8}{*}{LTMI} & ECI & 0.3045 & 0.3911 & 0.4509 & 8.8375 & 0.1810 & \textbf{0.1 min} \\
    & ES-MDA & \bestcell{0.1250} & \bestcell{0.2007} & \bestcell{0.2842} & \bestcell{2.4285} & \bestcell{0.0612} & \textbf{0.1 min} \\
    & FunDPS & 0.1611 & 0.2867 & 0.3327 & \secondcell{2.8501} & \secondcell{0.0640} & 5.4 min \\
    & Fun-DDPS & 0.1699 & 0.2374 & 0.3326 & 2.9389 & 0.1828 & 4.8 min \\
    & DiffusionPDE & 0.1593 & \secondcell{0.2265} & 0.3183 & 2.9462 & 0.1554 & 10.8 min \\
    & DDIS & 0.1504 & 0.2265 & \secondcell{0.3033} & 2.9999 & 0.1869 & 4.8 min \\
    & FunDiff & \secondcell{0.1492} & 0.3994 & 0.3753 & 3.1404 & 0.2624 & 0.4 min \\
    & MC-dropout & 0.1554 & 0.8542 & 0.6707 & 4.6755 & 0.5060 & \textbf{0.1 min} \\
    \bottomrule
  \end{tabular}
  }
\end{table}

\begin{figure}[H]
  \centering
  \includegraphics[width=0.82\textwidth]{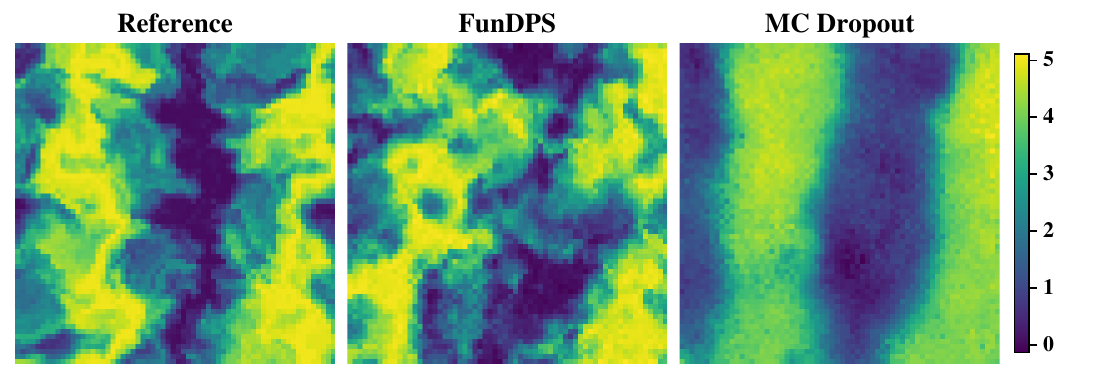}
  \caption{LTMI case visualization for the first material field $\sigma_{t1}$.
  The panels show one reference posterior sample, one FunDPS-generated sample, and one FNO with MC-Dropout-generated sample.
  Although MC Dropout attains lower LTMI posterior-mean error than FunDPS in \Cref{tab:main_results}, its sample is visibly over-smoothed relative to the reference structure, consistent with its high posterior-standard-deviation error, MMD, and SWD.
  }
  \label{fig:ltmi_case4_mcdropout_visualization}
\end{figure}

\textbf{Mean-only summaries can be misleading.}
A central purpose of PosteriorBench is to prevent posterior evaluation from collapsing back to a single point-summary comparison.
The LTMI results provide a concrete example: FNO with MC Dropout attains lower posterior-mean error than FunDPS in \Cref{tab:main_results}, yet its posterior-std error, MMD, and SWD remain high.
\Cref{fig:ltmi_case4_mcdropout_visualization} shows why this is not a metric inconsistency:
For a representative LTMI case, the MC Dropout sample is visibly over-smoothed relative to the reference posterior sample, while FunDPS better preserves the fine spatial texture of the material field.
Thus, a low mean error can reflect agreement with a central tendency while still missing the geometry and spread of the posterior distribution.
Marginal moment consistency should therefore be read jointly, together with metrics like MMD and SWD.

\textbf{Useful inductive bias in Function-space diffusion samplers.}
Across the benchmark, FunDPS, Fun-DDPS, and DDIS form a family of guided diffusion posterior samplers built on function-space score priors.
At the same time, their performance is task-dependent: classical methods such as ES-MDA can be stronger in settings such as LTMI, and the best-performing function-space diffusion sampler varies across datasets.
Their results suggest that function-space score priors are useful for posterior matching under sparse or low-resolution observations.
The comparison with DiffusionPDE points to the value of function-space score backbones, whose spectral operator blocks are better aligned with continuous physical fields than the grid-based backbone.
The comparison with ECI highlights the role of soft likelihood guidance: ECI enforces observed entries directly through hard replacement, whereas guided diffusion samplers condition the sampling process on the observations while trying to stay on the prior manifold.
During inference, these samplers dynamically balance learned prior against data consistency, which is central to recovering ensembles rather than only point reconstructions.
The remaining variation across tasks and metrics motivates a closer comparison within this guided diffusion family.

\begin{figure}[H]
  \centering
  \includegraphics[width=\textwidth]{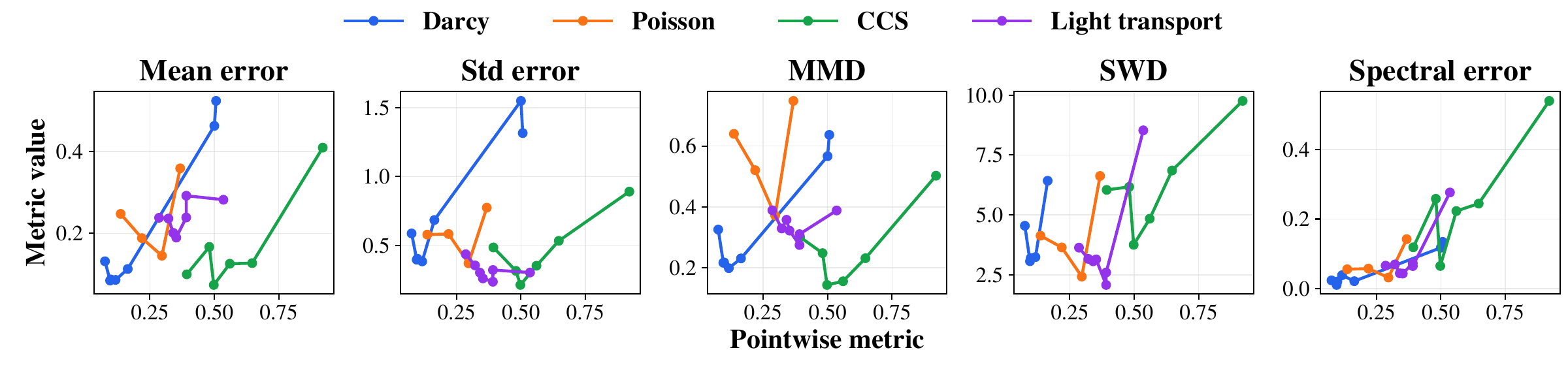}
  \vspace{-1em}
  \caption{Relationship between a traditional pointwise metric and the five posterior metrics.
  Each panel compares the pointwise metric with one posterior metric across methods and tasks.
  The V-shaped trends show that the distributional metrics are not monotone with respect to pointwise error.
  }
  \label{fig:guidance_sweep_pointwise_distributional}
\end{figure}

\subsection{Pointwise-Distributional Tradeoff}
In addition to the five posterior metrics used in \Cref{tab:main_results}, we compare a traditional pointwise relative $L^2$ metric against one reference sample.
\Cref{fig:guidance_sweep_pointwise_distributional} shows that the relationship is often V-shaped rather than monotone.
At the low-pointwise-error side, pushing samples closer to a single reference can remove distributional structure, so distribution-sensitive errors increase even as the pointwise metric improves.
At the high-error side, both pointwise and posterior errors are large, and the two families of metrics become positively associated.
This pattern indicates that a pointwise metric alone can misidentify over-fitted or over-concentrated samples.
The five posterior metrics are therefore useful because they expose the trade-off between single-reference reconstruction and distributional fidelity.

\subsection{Qualitative Posterior Comparisons}
\begin{figure}[!htbp]
  \centering
  \includegraphics[width=0.975\textwidth]{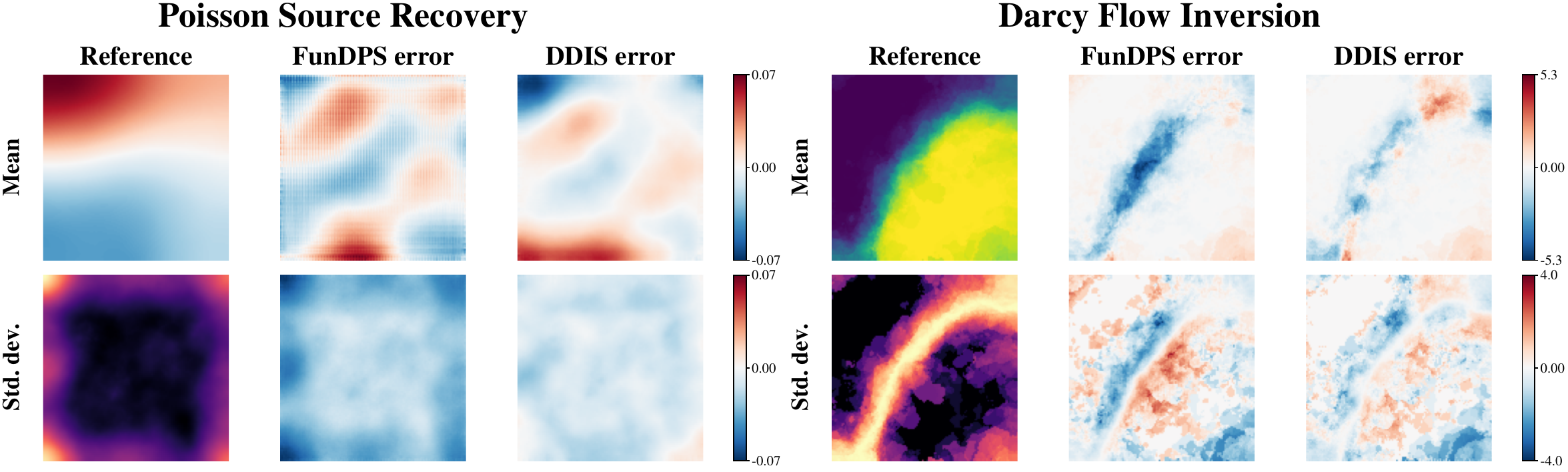}
  \caption{Qualitative posterior comparisons.
  Left: Poisson source recovery.
  Right: Darcy flow inversion.
  For each problem, the top row shows posterior mean $\mu$ and the bottom row shows posterior standard deviation $\sigma$.
  Within each group, we show the reference posterior statistic, FunDPS error, and DDIS error from left to right.
  }
  \label{fig:posterior_comparison}
\end{figure}

\Cref{fig:posterior_comparison} compares posterior mean and standard deviation estimates on Poisson source recovery and Darcy flow inversion.
Both FunDPS and DDIS recover the main posterior structures, while their error maps reveal larger discrepancies in high-gradient and high-uncertainty regions.
Both models also exhibit conservative predictions.
For Poisson source recovery, the mean error maps show residuals that pull extreme values toward zero, indicating that both models underestimate the magnitude of the source extrema.
This conservative behavior extends to uncertainty quantification: the standard-deviation error maps for both Poisson source recovery and Darcy flow inversion are predominantly negative, suggesting that the models systematically underestimate posterior variance.
Beyond these shared traits, errors in Darcy flow inversion concentrate near sharp interface-like structures, highlighting a more challenging posterior landscape.
Overall, DDIS produces more spatially balanced residuals and fewer large localized artifacts, though accurately capturing the full scale of posterior extremes and standard deviation remains a shared challenge.

\subsection{Out-of-Distribution Experiment}

We evaluate out-of-distribution generalization on Poisson source recovery by varying the GRF parameter range used to train the Fun-DDPS prior.
The prior is trained under three $\alpha$ regimes: \textit{single} ($\alpha\equiv2.25$), \textit{narrow} ($\alpha \in 2.25 \pm 0.375$), and \textit{full} ($\alpha\in 2.25 \pm 0.75$).
The differentiable surrogate used for likelihood guidance is trained on the \textit{full} parameter range in all three settings, so the ablation isolates the effect of prior-training coverage.
\Cref{tab:poisson_ood_funddps_metrics} reveals a counterintuitive split between pointwise and distributional behavior.
The \textit{full} prior gives the lowest pointwise relative $L^2$, but the distribution-sensitive metrics (std error, MMD, and SWD) worsen as the prior-training range expands.
Thus, broader prior coverage might not automatically improve posterior matching when the prior must represent source fields from a larger, harder-to-learn range.

\begin{figure}[htbp]
  \centering
  \includegraphics[width=\textwidth]{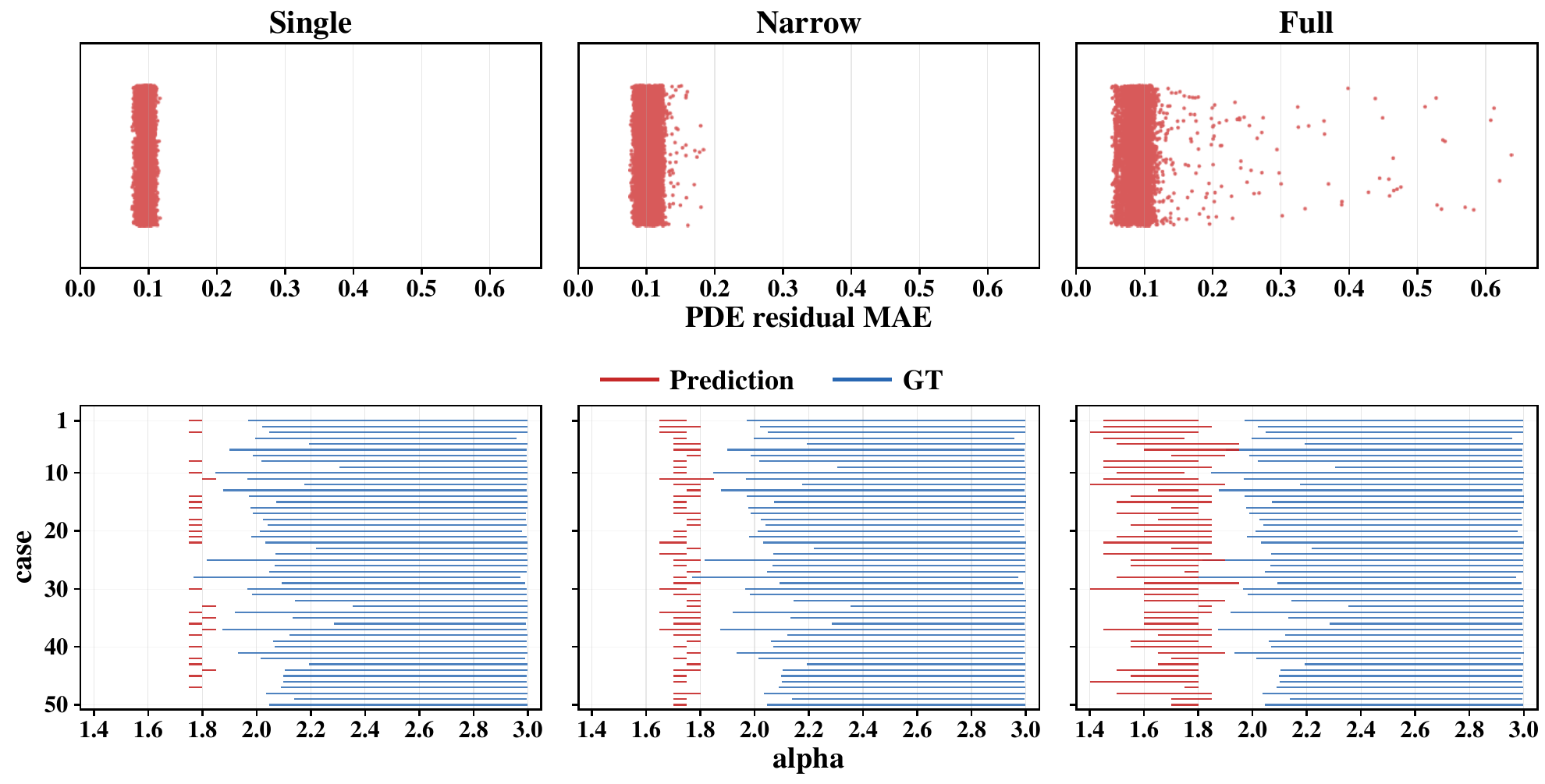}
  \caption{Fun-DDPS Poisson OOD diagnostics.
  Top row: PDE residual diagnostic, where each point corresponds to a generated sample evaluated by using the predicted pair $(\hat f,\hat \phi)$ and computing the residual $A(\hat\phi)-\hat f$ under the Poisson operator.
  Bottom row: $\alpha$-recovery diagnostic across single, narrow, and full prior-training ranges, where recovered $\alpha$ ranges are estimated from generated source fields using a DCT-domain GRF spectral likelihood and compared with the reference posterior ranges.}
  \label{fig:poisson_ood_pde_residual}
  \label{fig:poisson_ood_alpha_ranges}
\end{figure}

To diagnose this behavior, \Cref{fig:poisson_ood_pde_residual} evaluates physics and parameter consistency.
(i) We compute the Poisson residual $|A(\hat\phi)-\hat f|$ for each predicted pair $(\hat f,\hat\phi)$.
The \textit{full} setting produces more high-residual outliers, indicating more frequent PDE violations.
(ii) We estimate the GRF smoothness $\alpha$ from each generated source using a DCT-domain spectral likelihood.
The estimates are consistently biased below the reference posterior; broader training ranges widen their distribution mainly toward lower values, revealing poor recovery of latent smoothness despite plausible pixel-space samples.

Taken together, the PDE-residual and $\alpha$-recovery diagnostics further show that lower pointwise error need not imply better posterior distribution matching.
The distributional metrics align more closely with latent-parameter recovery and physical-consistency diagnostics than pointwise relative $L^2$ alone.
A second finding is that Poisson source recovery remains a demanding benchmark despite being generated from a synthetic GRF family.
Variation in the latent smoothness parameter induces a sufficiently complex distribution that strong samplers do not fully recover the parameter structure.

\begin{table}[htbp]
  \centering
  \caption{Fun-DDPS Poisson out-of-distribution evaluation across GRF prior-training ranges.
  Single uses $\alpha=2.25$, Narrow uses $\alpha\in[1.875,2.625]$, and Full uses $\alpha\in[1.5,3.0]$.}
  \label{tab:poisson_ood_funddps_metrics}
  \resizebox{\textwidth}{!}{%
  \begin{tabular}{lcccccc}
    \toprule
    Prior coverage & Pointwise Rel L2 & Mean Rel L2 & Std Rel L2 & MMD & SWD & Spectral Rel L2 \\
    \midrule
    Single & 0.4176 & 0.1551 & 0.3235 & 0.2669 & 1.7041 & 0.0330 \\
    Narrow & 0.3418 & 0.1487 & 0.3151 & 0.2879 & 1.8243 & 0.0358 \\
    Full & 0.2956 & 0.1464 & 0.3720 & 0.3730 & 2.4415 & 0.0329 \\
    \bottomrule
  \end{tabular}
  }
\end{table}

\subsection{Observation Noise and Guidance Calibration}
This experiment tests whether solver hyper-parameters that are often described as inverse noise scales actually calibrate to the posterior induced by different observation-noise levels.
As shown in \Cref{eq:gaussian_observation_model,eq:gaussian_likelihood_score}, the Gaussian likelihood score scales with the likelihood precision $\sigma_y^{-2}$, which motivates scaling data-consistency guidance with the observation-noise level.
However, the guidance coefficient used in practice may not follow this rule due to the interaction with discretization, normalization, annealing schedules, etc.

\begin{figure}[h]
  \centering
  \includegraphics[width=0.7\textwidth]{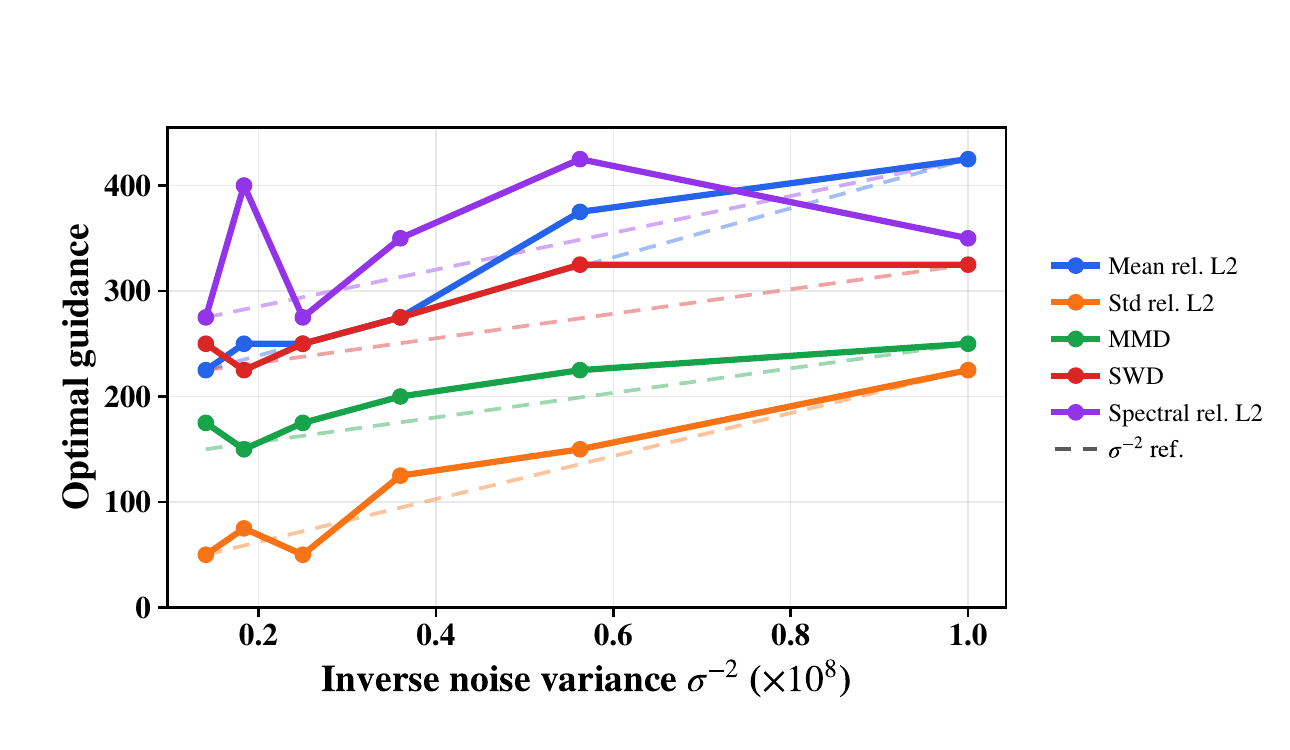}
  \vspace{-1em}
  \caption{Relationship between inverse observation-noise variance $\sigma^{-2}$ and the optimal guidance weight for FunDPS on Poisson source recovery, with posterior threshold set to $\epsilon=3\sigma$.
  Each curve reports the guidance value that minimizes one evaluation metric at each inverse noise variance.
  The optimal guidance weights generally increase with $\sigma^{-2}$, indicating a positive association between likelihood precision and preferred guidance strength.
  The metric-specific optima also differ at the same $\sigma^{-2}$, notably between posterior-mean and posterior-standard-deviation errors.
  }
  \label{fig:noise_guidance_ablation}
\end{figure}

For Poisson source recovery, we vary the observation-noise scale $\sigma$ and report the guidance optimum $\lambda_m^\star(\sigma)=\arg\min_{\lambda} m(\mathcal{P}_{gen}^{\sigma,\lambda},\mathcal{P}_{ref}^{\sigma})$.
\Cref{fig:noise_guidance_ablation} shows that the optimal guidance weight generally increases with the inverse observation-noise variance $\sigma^{-2}$ for almost all metrics.
Spectral error exhibits larger fluctuations, but still follows the same broad positive relation.
This behavior is consistent with the Gaussian likelihood assumption.
The full guidance-weight sweep and case-level spatial diagnostics are reported in \Cref{app:poisson_guidance_calibration}.

The same sweep also shows that the optimum is not metric-invariant.
At a fixed noise scale $\sigma$, the values of $\lambda_m^\star(\sigma)$ are different across metrics, especially between posterior-mean and posterior-std errors.
Because the mean and standard deviation are complementary summaries of the same posterior distribution, this separation indicates that the guidance strength that best fits posterior center need not be variance-calibrated.
Thus, a single scalar guidance weight faces trade off between mean accuracy and uncertainty calibration.

\subsection{Resolution Ablation}
The resolution ablation study in \Cref{tab:resolution_ablation} shows that FunDPS consistently outperforms DiffusionPDE across all training settings when evaluated at $128 \times 128$.
FunDPS achieves lower mean and standard-deviation errors, as well as improved MMD, SWD, and spectral metrics, with the best distributional alignment under multi-resolution training.
In contrast, DiffusionPDE shows limited gains with increased resolution.
These results indicate that function-space training improves robustness to train-test resolution changes.

\begin{table}[h]
  \centering
  \caption{Resolution generalization study on Darcy flow inversion.
  The table compares DiffusionPDE and FunDPS at each training resolution setting.
  The inference is conducted at $128 \times 128$ resolution.}
  \label{tab:resolution_ablation}
  \resizebox{\textwidth}{!}{%
  \begin{tabular}{llccccc}
    \toprule
    Resolution & Method & Mean error $\downarrow$ & Std error $\downarrow$ & MMD $\downarrow$ & SWD $\downarrow$ & Spectral error $\downarrow$ \\
    \midrule
    \multirow{2}{*}{$64 \times 64$} & DiffusionPDE & 0.1838 $\pm$ 0.0433 & 0.9324 $\pm$ 0.1902 & 0.3361 $\pm$ 0.0506 & 1.6860 $\pm$ 0.4621 & 0.0690 $\pm$ 0.0478 \\
    & FunDPS & \textbf{0.1070} $\pm$ 0.0191 & \textbf{0.4540} $\pm$ 0.0935 & \textbf{0.2570} $\pm$ 0.0474 & \textbf{0.8440} $\pm$ 0.2320 & \textbf{0.0250} $\pm$ 0.0179 \\
    \midrule
    \multirow{2}{*}{$128 \times 128$} & DiffusionPDE & 0.1745 $\pm$ 0.0319 & 0.6708 $\pm$ 0.1016 & 0.3306 $\pm$ 0.0425 & 1.2667 $\pm$ 0.4895 & 0.0506 $\pm$ 0.0348 \\
    & FunDPS & \textbf{0.0926} $\pm$ 0.0182 & \textbf{0.3570} $\pm$ 0.0396 & \textbf{0.2030} $\pm$ 0.0292 & \textbf{0.8090} $\pm$ 0.2820 & \textbf{0.0158} $\pm$ 0.0129 \\
    \midrule
    \multirow{2}{*}{Multi-res} & DiffusionPDE & 0.1833 $\pm$ 0.0423 & 0.9324 $\pm$ 0.1860 & 0.3368 $\pm$ 0.0494 & 1.5465 $\pm$ 0.5450 & 0.0661 $\pm$ 0.0485 \\
    & FunDPS & \textbf{0.0886} $\pm$ 0.0237 & \textbf{0.4010} $\pm$ 0.0821 & \textbf{0.2210} $\pm$ 0.0451 & \textbf{0.7400} $\pm$ 0.1980 & \textbf{0.0154} $\pm$ 0.0110 \\
    \bottomrule
  \end{tabular}
  }
\end{table}

\section{Discussion}
\label{sec:discussion}

PosteriorBench evaluates generative scientific inverse solvers as posterior samplers rather than as single-reconstruction methods.
Across Darcy flow inversion, Poisson source recovery, carbon capture and storage, and light-transport material inference, the benchmark makes the residual ambiguity under partial observations explicit by comparing generated ensembles against reference posterior distributions.
The experiments highlight two main findings.
First, PosteriorBench exposes concrete regimes in which pointwise and distributional evaluation disagree.
Second, function-space diffusion samplers provide strong posterior-matching performance across the benchmark, yet still leave gaps in jointly matching posterior means and standard deviations, recovering latent parameters, and satisfying PDE constraints.
These findings highlight that a solver can match observations or point estimates while still misrepresenting posterior uncertainty, spatial structure, or distinct meaningful modes.

The present benchmark is necessarily limited by the cost of constructing high-fidelity reference posteriors and the number of solvers evaluated so far.
Our broader vision is for PosteriorBench to serve as an evolving evaluation platform expanding through community effort, while ultimately shifting the field away from single point reconstructions and toward calibrated, reproducible posterior recovery for scientific decision making under uncertainty.

\begin{ack}
Anima Anandkumar is supported in part by Bren endowed chair, ONR (MURI grant N00014-23-1-2654), and the AI2050 senior fellow program at Schmidt Sciences. Jiachen Yao is supported in part by the Naren and Vinita Gupta Fellowship.
The authors thank Modal for providing part of the compute credits.
\end{ack}

\bibliographystyle{unsrt}
\bibliography{references}

\clearpage
\appendix

\section{Related Work}
\label{app:related_work}

\paragraph{Bayesian scientific inverse problems.}
Inverse problems are classically formulated as the recovery of unknown parameters or fields from indirect observations through a forward physical model~\citep{tarantola2005inverse,groetsch1993inverse,beck1985inverse,mueller2012linear}.
The Bayesian formulation treats the unknown as a random function and combines a prior with the observation likelihood to obtain a posterior distribution~\citep{cotter2009bayesian,gelman1995bayesian}.
This view is essential when observations are sparse, noisy, or non-identifying, since posterior uncertainty can remain large even when the observations are matched.
Classical data-assimilation and ensemble methods, including ensemble Kalman approaches, provide scalable approximations for some high-dimensional systems but can be limited by Gaussian or linearized update assumptions~\citep{iglesias2013ensemble}.
In high-dimensional scientific settings, asymptotically reliable samplers such as MCMC, sequential Monte Carlo, or rejection sampling are often too expensive to use as practical solvers, which motivates amortized or generative approximations~\citep{cardoso2023monte,dou2024diffusion}.
PosteriorBench uses these slow methods not as deployment algorithms, but as reference procedures for evaluating whether faster learned samplers recover the intended posterior.

\paragraph{Neural operators and PDE inverse solvers.}
Physics-informed neural networks and neural operators have become standard tools for learning PDE solution maps and solving PDE-constrained inverse problems~\citep{raissi2019physics,li2020fourier,kovachki2023neural}.
Operator-learning architectures such as FNO and DeepONet learn mappings between function spaces and can generalize across discretizations more naturally than fixed-grid networks~\citep{li2020fourier,lu2019deeponet,kovachki2023neural}.
Subsequent architectures extend this idea with multipole, Laplace, transformer, and geometry-aware operator designs~\citep{li2020multipole,cao2024laplace,li2022transformer,li2023fourier}.
Physics-informed neural operators further add PDE residuals or weak supervision to improve generalization when paired data are limited~\citep{li2024physics}.
For inverse problems, deterministic neural solvers can be accurate when the target is a point estimate, but they do not by themselves represent the full posterior over plausible fields.
For example, neural inverse operators directly learn maps from observations to unknown coefficients~\citep{molinaro2023neural}.
PosteriorBench treats these models both as components of generative solvers and as important baselines or surrogates.

\paragraph{Uncertainty-aware neural samplers.}
Uncertainty-aware neural predictors provide another route to posterior sampling without training a full generative inverse model.
Bayesian neural networks place distributions over network weights and infer a weight posterior, so predictive variation reflects model uncertainty~\citep{mackay1992practical}.
MC dropout offers a scalable approximation by keeping dropout active at inference time and using repeated stochastic forward passes as approximate Bayesian predictions~\citep{gal2016dropout}.
In scientific machine learning, Bayesian PDE solvers such as B-PINNs~\citep{yang2021bpinns} and B-DeepONet~\citep{lin2023bdeeponet} combine Bayesian neural networks or Bayesian operator learning with PDE constraints to quantify uncertainty in forward and inverse settings.
These approaches are complementary to PosteriorBench, but their inverse examples typically recover single- or low-dimensional quantities, whereas PosteriorBench asks solvers to sample posteriors over $64{\times}64$ or $128{\times}128$ fields.
This high dimensionality is one reason generative inverse samplers are attractive, and hence why the benchmark focuses on them.
B-PINNs also require fitting a separate model for each observation case, which does not scale to the amortized multi-case evaluation used here.

\paragraph{Diffusion posterior sampling.}
Diffusion and score-based models were first developed as powerful unconditional generators~\citep{sohl2015deep,ho2020denoising,song2020score}, then adapted to inverse problems by conditioning a learned prior on measurements.
Conditional models learn task-specific conditional distributions~\citep{saharia2022palette,tashiro2021csdi}, whereas plug-and-play posterior samplers reuse an unconditional prior with an observation model at test time.
For linear or image-domain inverse problems, methods such as DDRM, DDNM, pseudoinverse, loss guidance, and RED-diff provide different mechanisms for combining denoising priors with data consistency~\citep{kawar2022denoising,wang2022zero,song2023pseudoinverse,song2023loss,wu2024principled,mardani2023variational}.
Sequential Monte Carlo and filtering perspectives seek stronger posterior correctness guarantees for some classes of inverse problems~\citep{cardoso2023monte,dou2024diffusion}.
DAPS reduces approximation error by decoupling denoising and likelihood updates~\citep{zhang2025improving}.
These advances motivate distributional evaluation, but many reported results still emphasize reconstruction quality, perceptual quality, or observation consistency rather than calibrated posterior matching.

\paragraph{Diffusion models for scientific applications.}
Scientific inverse problems add challenges that are muted in natural-image restoration: the forward map may be a PDE solver, observations may live in a different physical field than the unknown, and the state is more naturally a function than a fixed-resolution image.
DiffusionPDE models paired physical fields under partial observation~\citep{huang2024diffusionpde}; physics-informed diffusion methods add residual or constraint guidance~\citep{shu2023physics,bastek2025physicsinformeddiffusionmodels,zhang2025physicsinformeddistillationdiffusionmodels}; and CoCoGen constructs conditioned score-based models for forward and inverse physical problems~\citep{jacobsen2025cocogen}.
Function-space approaches such as FunDPS and FunDiff address discretization dependence by defining generative modeling over functions rather than only arrays~\citep{yao2025guided,wang2025fundiffdiffusionmodelsfunction}.
DDIS and related decoupled designs separate prior learning from the physics-induced likelihood using a neural operator surrogate~\citep{lin2026decoupled}.
Other recent work explores diffusion bridges, wavelet diffusion operators, sparse flow-field reconstruction, and conditional PDE simulation~\citep{li2025physicsaligned,hu2025waveletdiffusionneuraloperator,Amor_s_Trepat_2026,shysheya2024conditional}.
PosteriorBench provides a common setting for comparing these design choices as posterior samplers.

\paragraph{Benchmarks for scientific inverse problems.}
Scientific inverse-problem benchmarks increasingly extend beyond natural-image restoration.
InverseBench tests plug-and-play diffusion priors with physical forward models, but is limited to single-reference reconstruction~\citep{zheng2025inversebench}.
Simulation-based inference benchmarks more directly assess posterior estimation, but often consider lower-dimensional parameters and tractable likelihoods~\citep{lueckmann2021benchmarking}; learned posterior targets can also make scores depend on model architecture, training coverage, and calibration~\citep{wehenkel2025addressing}.
Zach et al. enable precise distributional tests for one-dimensional Bayesian linear inverse problems with 64 discretization points and L\'{e}vy-process priors, whose posteriors admit efficient Gibbs sampling~\citep{zach2025statistical}.
PosteriorBench instead targets scientific function-space posteriors on $64{\times}64$ or $128{\times}128$ fields under sparse, low-resolution, or column observations generated by expensive physics.
It constructs reference distributions through prior sampling, physical simulation, observation matching, and importance weighting, enabling posterior-level evaluation across multiple physics domains, including CCS and LTMI.

\paragraph{Carbon capture and storage.}
Carbon capture and storage is a major climate-mitigation technology, but safe deployment requires uncertainty-aware monitoring of subsurface CO$_2$ migration and pressure buildup~\citep{pacala2004stabilization}.
Reservoir characterization and data assimilation have long relied on ensemble Kalman methods and ensemble smoothers, including ES-MDA, because they can update high-dimensional geomodel ensembles from sparse monitoring data~\citep{Emerick2013b,Jung2018}.
These methods are practical and domain-relevant but can struggle with strongly non-Gaussian geological priors and channelized or facies-like structures.
Deep generative parameterizations and learned geomodel priors have been explored as a way to represent non-Gaussian reservoir structure within data-assimilation workflows~\citep{zhu2018bayesian,di2025latent}.
Neural-operator surrogates have accelerated geological CO$_2$ storage simulation~\citep{wen2022u,wen2023real}, and recent work combines generative models with data assimilation, history matching, or inverse modeling for geological carbon storage~\citep{Seabra2024GCSDA,jiang2024history,han2024surrogate,teng2025likelihood,wang2025generative,feng2025generative}.
PosteriorBench includes CCS to test posterior recovery in a realistic sparse-well setting, with ESMDA retained as a CCS-specific baseline.

\section{Data Generation Details}
\label{app:data_generation}

\subsection{Rejection sampling}
\label{sec:appendix_rejection_sampling}

To evaluate posterior-generating inverse solvers, we establish high-quality reference posterior samples using an accelerated rejection sampling scheme.
Since querying the numerical forward solver sequentially during inference is computationally prohibitive, particularly for complex physical systems, we adopt an offline-to-online sampling strategy that leverages spatial invariances.

\paragraph{Offline Prior Pool Generation.}
We first construct a massive offline dataset, denoted as the prior pool $\mathcal{D}_{\text{pool}} = \{ (\mathbf{x}^{(i)}, \mathbf{y}^{(i)}) \}_{i=1}^{N}$, where $N$ denotes a sufficiently large, task-specific pool size. Here, $\mathbf{x}^{(i)} \sim p(\mathbf{x})$ represents a generalized input physical parameter field (e.g., permeability, forcing terms, scattering coefficients, or geological models) sampled from the prior distribution, and $\mathbf{y}^{(i)} = \mathcal{F}(\mathbf{x}^{(i)})$ is the corresponding physical system state obtained via the forward numerical solver $\mathcal{F}$.

The configuration of the prior, the numerical solver $\mathcal{F}$, and the computational backend depend heavily on the specific partial differential equation (PDE) task:
\begin{itemize}
    \item \textbf{Darcy Flow and Poisson Equation (JAX-accelerated):} The input priors are constructed based on Gaussian Random Fields (GRFs). For Darcy flow, a binary field is generated by thresholding a GRF; for the Poisson equation, continuous GRFs with varying length scales and smoothness parameters ($\tau, \alpha$) are used to ensure a diverse distribution of spatial frequencies. Because these setups allow for highly parallelizable synthetic generation, both solvers are explicitly optimized using JAX, enabling massive batched execution on GPUs.
    
    \item \textbf{Light Transport and Carbon Capture and Storage (CCS):} Unlike strictly synthetic GRF setups, these tasks involve highly complex physical structures (such as varying scattering media for light transport and realistic geomodels for CCS). The respective prior pools are generated using their domain-specific, high-fidelity physical simulators, which are necessary to accurately capture the complex, non-linear forward dynamics governing light scattering or multiphase fluid flow.
\end{itemize}

\paragraph{Observation Operators ($\mathcal{H}$).}
During the online sampling phase, we define a target pair $(\mathbf{x}_{\text{gt}}, \mathbf{y}_{\text{gt}})$ and obtain simulated observations $o_{\text{gt}} = \mathcal{H}(\mathbf{y}_{\text{gt}})$.
We denote $i$ and $j$ as the discrete spatial indices across the computational grid.
To test the benchmark solvers across measurement regimes, we consider three distinct observation operators $\mathcal{H}$ tailored to different physical scenarios:
\begin{enumerate}
    \item \textit{Sparse Random Observation (Darcy, Poisson):} Sensors are randomly scattered across the spatial domain. The operator is defined as $\mathcal{H}_{\text{sparse}}(\mathbf{y}) = \{ \mathbf{y}(i_m, j_m) \}_{m=1}^{M}$, where $M$ is the total number of sensors and $(i_m, j_m)$ denotes the discrete spatial coordinates of the $m$-th.
    \item \textit{Low-Resolution Observation (Darcy, Poisson, Light Transport):} The system provides a coarse-grained view of the full solution, modeled via an average pooling operation: $\mathcal{H}_{\text{low-res}}(\mathbf{y}) = \text{AvgPool}(\mathbf{y}, s)$, which downsamples the original high-resolution field to a coarser grid of scale $s \times s$ (e.g., $16 \times 16$).
    \item \textit{Column Observation (CCS):} Consistent with well-log data in geophysics, observations are only available along specific vertical columns. The operator extracts data strictly along these vertical indices: $\mathcal{H}_{\text{col}}(\mathbf{y}) = \{ \mathbf{y}(i_c, j) \mid \forall j, c \in \mathcal{C} \}$, where $\mathcal{C}$ is the set of observable column indices $i_c$.
\end{enumerate}

\paragraph{Rejection Sampling and Importance Weighting.}
To efficiently obtain posterior samples from $\mathcal{D}_{\text{pool}}$ given $o_{\text{gt}}$, we employ a rejection sampling scheme based on the observation discrepancy. For tasks defined on a Cartesian grid with isotropic physical properties, specifically Darcy Flow and the Poisson Equation, we further exploit their discrete rotational equivariance. For these tasks, we apply discrete spatial rotations $\mathcal{R}_k \in \{0^\circ, 90^\circ, 180^\circ, 270^\circ\}$ to each candidate pair $(\mathbf{x}^{(i)}, \mathbf{y}^{(i)})$, effectively quadrupling the pool size without additional solver calls. In contrast, for Light Transport and CCS, samples are utilized in their original orientation to preserve task-specific physical constraints.

A candidate $(\mathbf{x}^{(i, k)}, \mathbf{y}^{(i, k)})$ is accepted if the Root Mean Square Error (RMSE) between its observation and the target observation falls below a predefined threshold $\epsilon$:
\begin{equation}
    \text{RMSE}^{(i, k)} =
    \left(
        \frac{1}{|\Omega_{\text{obs}}|}
        \left\| \mathcal{H}(\mathbf{y}^{(i, k)}) - o_{\text{gt}} \right\|_2^2
    \right)^{1/2}
    \le \epsilon
\end{equation}
where $|\Omega_{\text{obs}}|$ denotes the total number of observation points. We continue the search until $K$ valid posterior samples (e.g., $K=100$) are collected for the given target. 

Finally, to account for the continuous nature of the posterior probability, we assign an importance weight $w^{(i,k)}$ to each accepted sample based on a Gaussian likelihood formulation. The unnormalized weights are computed as:
\begin{equation}
    \tilde{w}^{(i,k)} = \exp \left( -\frac{1}{2\sigma^2} \left( \text{RMSE}^{(i, k)} \right)^2 \right)
\end{equation}
To ensure the likelihood properly reflects the acceptance criterion, we set the standard deviation to $\sigma = \frac{1}{3} \epsilon$, ensuring that samples near the acceptance boundary $\epsilon$ are appropriately down-weighted. The final weights are normalized such that $\sum \tilde{w}^{(i,k)} = 1$, yielding a weighted empirical posterior distribution that rigorously approximates the true Bayesian posterior.

\begin{table}[htbp]
  \centering
  \caption{Rejection-sampling hyperparameters for Darcy flow and Poisson source recovery.}
  \label{tab:rejection_sampling_params}
  \begin{tabular}{@{}lcc@{}}
    \toprule
    \textbf{Parameter} & \textbf{Darcy flow} & \textbf{Poisson} \\
    \midrule
    \multicolumn{3}{@{}l}{\textit{Shared settings}} \\
    \quad Resolution & \multicolumn{2}{c}{128} \\
    \quad Threshold ($\eta$) & \multicolumn{2}{c}{$4 \times 10^{-4}$} \\
    \quad Noise ($\sigma$) & \multicolumn{2}{c}{$1.33 \times 10^{-4}$} \\
    \quad Random sensors & \multicolumn{2}{c}{128} \\
    \addlinespace[2pt]
    \multicolumn{3}{@{}l}{\textit{Prior specification}} \\
    \quad Smoothness ($\alpha$) & 2.0 & $1.5 \sim 3.0$ \\
    \quad Correlation length ($\tau$) & 3.0 & $2.0 \sim 4.0$ \\
    \bottomrule
  \end{tabular}
\end{table}

\subsection{Carbon Capture and Storage}
\label{app:ccs_setup}

The CCS benchmark uses a synthetic dataset of supercritical CO$_2$ injection into a radially symmetric deep saline aquifer.
Each realization pairs a heterogeneous permeability field $m \in \mathbb{R}^{64 \times 200}$ with the corresponding CO$_2$ saturation field $s = F(m) \in \mathbb{R}^{64 \times 200}$ after 30 years of continuous injection.

\paragraph{Governing equations.}
The forward model solves the conservation of mass for a two-phase (CO$_2$--water) system in porous media.
For phase $\alpha \in \{w, g\}$ (water and gas), the mass balance reads
\begin{equation}
    \frac{\partial}{\partial t}(\phi\,\rho_\alpha S_\alpha)
    + \nabla\cdot(\rho_\alpha\,\mathbf{u}_\alpha)
    = q_\alpha,
\end{equation}
where $\phi$ is porosity, $\rho_\alpha$ is density, $S_\alpha$ is saturation, and $\mathbf{u}_\alpha$ is the Darcy velocity
\begin{equation}
    \mathbf{u}_\alpha
    = -\frac{k_{r\alpha}\,\mathbf{K}}{\mu_\alpha}
      (\nabla P_\alpha - \rho_\alpha\,\mathbf{g}).
\end{equation}
Here $\mathbf{K}$ is the absolute permeability tensor (the unknown field), $k_{r\alpha}$ is relative permeability, $\mu_\alpha$ is viscosity, and $P_\alpha$ is phase pressure.
The system is closed by the saturation constraint $S_w + S_g = 1$ and the capillary pressure relation $P_c = P_g - P_w$.

\paragraph{Simulation domain.}
The domain represents an infinite-acting aquifer with a radius of 100\,km and a thickness of 135\,m, discretized on a 2D radial grid (64 depth $\times$ 200 radial cells).
No-flow conditions are enforced at the top and bottom caprock boundaries.
CO$_2$ is injected at a constant rate of 0.36\,Mt/year for 30 years through a single vertical well.

\paragraph{Geomodel prior.}
Permeability realizations are generated from geostatistical priors using sequential Gaussian simulation (SGeMS)~\citep{sgems}.
The geostatistical hyperparameters (mean permeability $\mu_{k_r} \sim \mathcal{U}[10, 500]$\,mD, standard deviation $\sigma_{k_r} \sim \mathcal{U}[1, 500]$\,mD, radial correlation length $\sim \mathcal{U}[359, 35{,}900]$\,m, and vertical correlation length $\sim \mathcal{U}[14, 56]$\,m) are sampled independently for each realization, creating a diverse prior that spans a wide range of geological scenarios.
The forward simulations are performed with the industry-standard reservoir simulator ECLIPSE (e300)~\citep{eclipse}.
The dataset comprises 12{,}000 training pairs and 1{,}390 test pairs.
Further details are provided by \cite{ju2026funddps}.

\paragraph{Observation Pattern}
Observations mimic realistic well-monitoring data: vertical column measurements are collected at two locations corresponding to an injection well ($x = 0$) and a monitoring well ($x = 50$, approximately 491\,m away).
Each column provides 64 saturation measurements along the depth axis, yielding 128 observed values out of 12{,}800 total grid cells ($1\%$ spatial coverage).
Gaussian observation noise with $\sigma_{\mathrm{obs}} = 0.04$ is added to the ground-truth saturation values at the observed locations.

\paragraph{Reference Posterior Construction}
The reference posterior for each test case is constructed via rejection sampling from a pool of 2 million prior geomodel samples.
Each candidate is evaluated through a pre-trained Local Neural Operator (LNO) surrogate~\citep{liu2024neural} that maps permeability to saturation, and the observation mismatch is computed at the monitored well locations.
Candidates are accepted with probability proportional to the Gaussian likelihood:
\begin{equation}
    L(m) = \exp\!\left(-\frac{\|M_{\mathrm{obs}} \odot (\mathcal{L}_\phi(m) - y_{\mathrm{obs}})\|^2}{2\sigma_{\mathrm{obs}}^2 \cdot |M_{\mathrm{obs}}|}\right),
\end{equation}
where $M_{\mathrm{obs}}$ is the binary observation mask and $|M_{\mathrm{obs}}|$ is the number of observed points.
This procedure yields approximately 26{,}000 accepted samples per case (acceptance rate ${\sim}1.3\%$), providing a dense empirical approximation to the true posterior.
The use of the neural-operator surrogate rather than the full simulator makes this large-scale rejection sampling computationally feasible.

\section{Reference Posterior Validation}
\label{app:reference_strength_validation}

Each metric is an estimator evaluated against a finite reference ensemble and therefore carries a variance component attributable to the reference set alone.
We isolate this component by constructing two independent reference ensembles, $\mathrm{GT}_1$ and $\mathrm{GT}_2$, at equal sampling budget and scoring the same solver outputs against each ensemble.
Under reference convergence, $\mathrm{GT}_1$ and $\mathrm{GT}_2$ are exchangeable and should agree up to Monte Carlo error.
For each metric $m$, we compute
\begin{equation}
  r_m =
  \max\left(
    \frac{m_{\mathrm{GT1}}}{m_{\mathrm{GT2}}},
    \frac{m_{\mathrm{GT2}}}{m_{\mathrm{GT1}}}
  \right).
\end{equation}
We report the worst-case ratio over all evaluated solvers in \Cref{tab:reference_ensemble_consistency}.
A value close to 1 indicates that the metric is insensitive to the particular finite reference ensemble used.

\begin{table}[htbp]
  \centering
  \caption{Reference-ensemble consistency check.
  Each entry reports the worst-case ratio $r_m$ over evaluated solvers when the same solver outputs are scored against two independent reference ensembles.}
  \label{tab:reference_ensemble_consistency}
  \resizebox{\textwidth}{!}{%
  \begin{tabular}{lccccc}
    \toprule
    Task & Mean Rel. L2 & Std Rel. L2 & MMD & SWD & Spectral Rel. L2 \\
    \midrule
    Poisson & 1.0295 & 1.0198 & 1.0112 & 1.0431 & 1.0674 \\
    Darcy Flow & 1.0272 & 1.0269 & 1.0196 & 1.0490 & 1.1146 \\
    Light Transport & 1.0132 & 1.0141 & 1.0090 & 1.0302 & 1.0589 \\
    Carbon Capture and Storage & 1.0767 & 1.0505 & 1.0366 & 1.0738 & 1.1266 \\
    \bottomrule
  \end{tabular}
  }
\end{table}

All ratios are close to 1, and all non-spectral metrics remain below 1.08.
These values are worst-case ratios rather than average-case ratios, making the check conservative.
The results suggest that the reference ensembles are sufficiently converged for the reported solver comparisons and make the empirical evaluation noise floor transparent.

\section{Additional Results}
\label{app:additional_results}

\subsection{Guidance Weight Calibration}
\label{app:poisson_guidance_calibration}

\begin{figure}[t]
  \centering
  \begin{minipage}{0.49\textwidth}
    \centering
    \includegraphics[width=\linewidth]{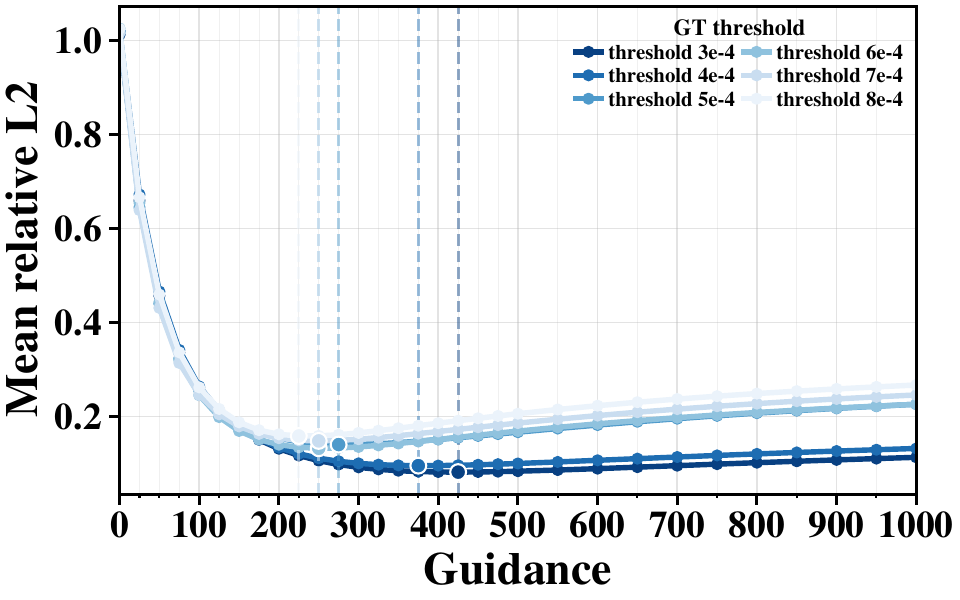}
    {\footnotesize (a) Posterior mean error}
  \end{minipage}\hfill
  \begin{minipage}{0.49\textwidth}
    \centering
    \includegraphics[width=\linewidth]{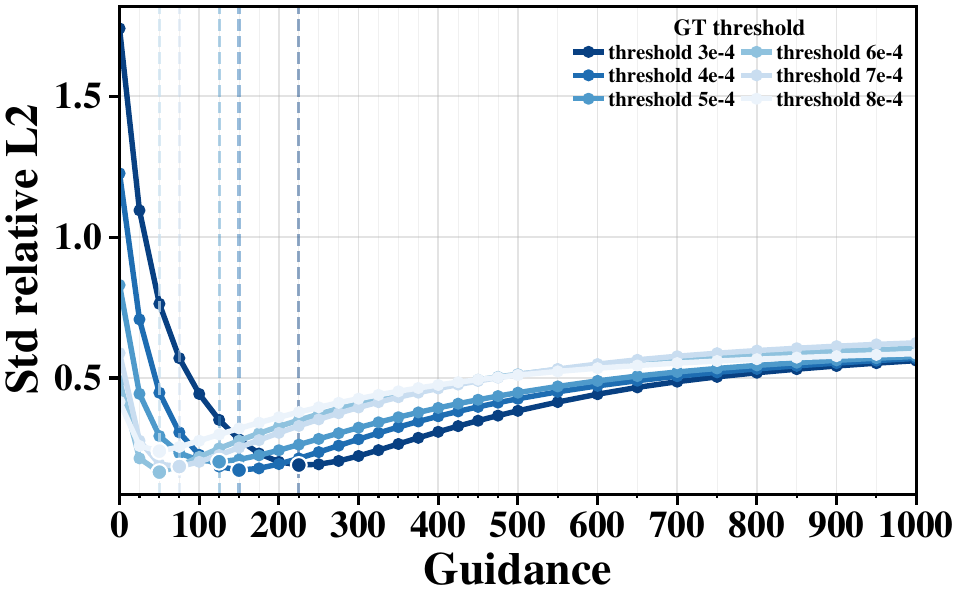}
    {\footnotesize (b) Posterior standard-deviation error}
  \end{minipage}

  \vspace{1.5mm}
  \begin{minipage}{0.49\textwidth}
    \centering
    \includegraphics[width=\linewidth]{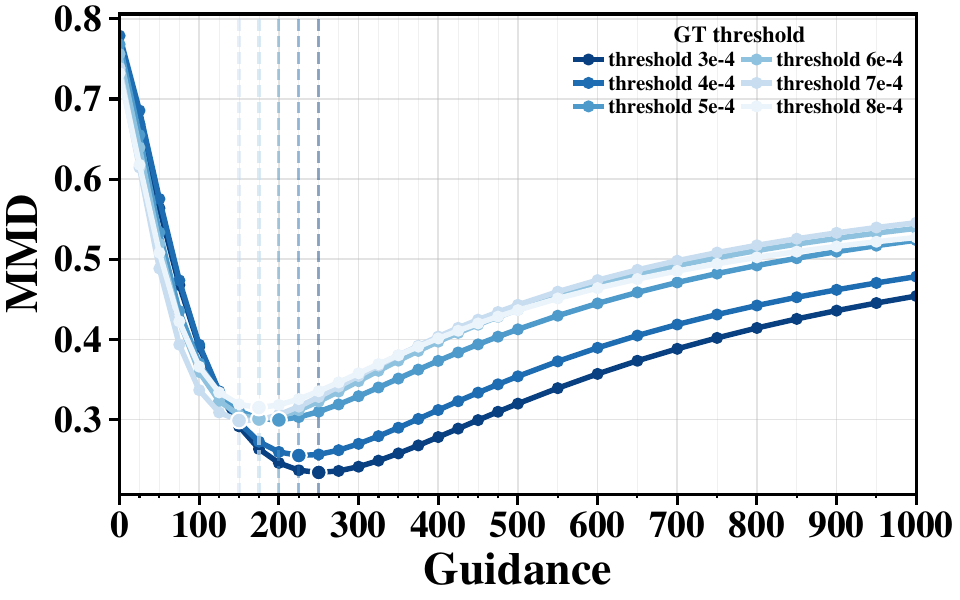}
    {\footnotesize (c) MMD}
  \end{minipage}\hfill
  \begin{minipage}{0.49\textwidth}
    \centering
    \includegraphics[width=\linewidth]{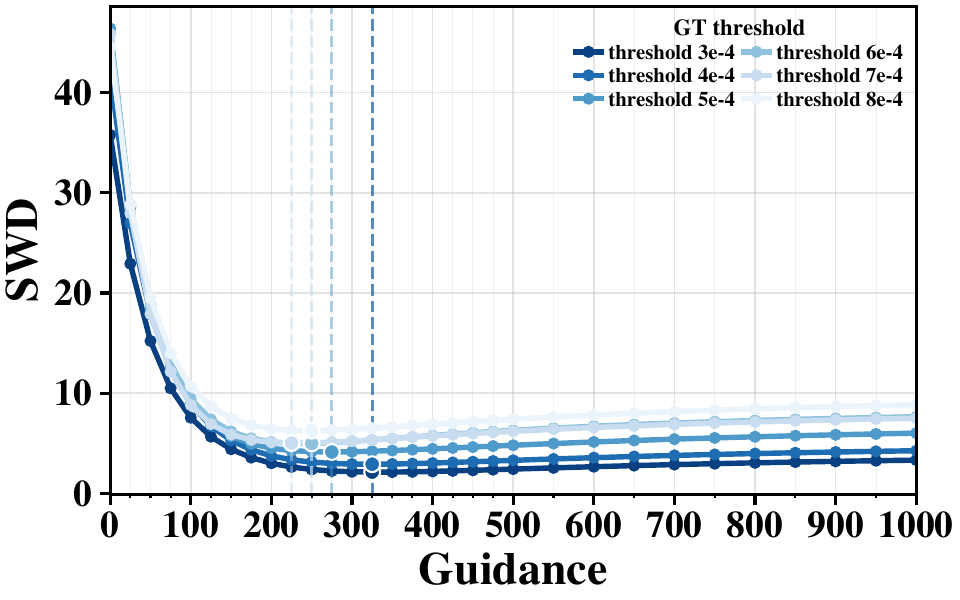}
    {\footnotesize (d) SWD}
  \end{minipage}

  \vspace{1.5mm}
  \begin{minipage}{0.58\textwidth}
    \centering
    \includegraphics[width=\linewidth]{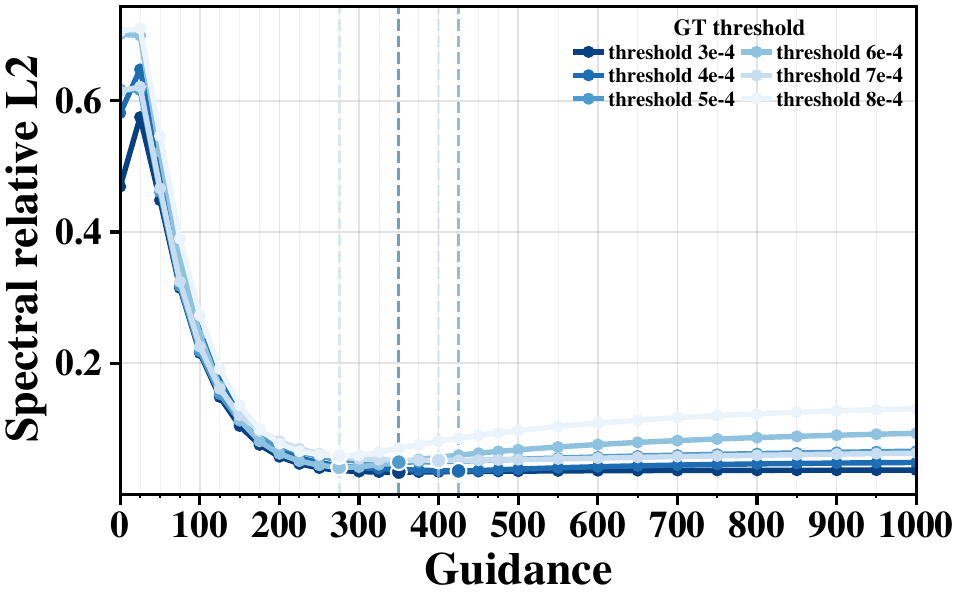}
    {\footnotesize (e) Spectral error}
  \end{minipage}
  \caption{Metric-specific guidance-weight sweeps for FunDPS on Poisson source recovery.
  Each panel plots one evaluation metric as a function of the guidance weight $\lambda$, with separate curves for posterior thresholds $\epsilon \in \{3,4,5,6,7,8\}\times 10^{-4}$.
  The curves expose both the broad decrease in preferred guidance strength as observation noise increases and the disagreement between metric-specific optima at the same threshold.}
  \label{fig:poisson_guidance_sweep_lines}
\end{figure}

\Cref{fig:poisson_guidance_sweep_lines} reports the full guidance-weight sweep for FunDPS on Poisson source recovery across posterior thresholds $\epsilon \in \{3,4,5,6,7,8\}\times 10^{-4}$ and guidance weights $\lambda \in [0,1000]$.
Each panel visualizes one posterior metric as a function of $\lambda$ under each threshold.
Across metrics, very small guidance weights under-condition on the observation, whereas overly large weights increasingly distort posterior calibration.
The location of the optimum generally shifts toward smaller $\lambda$ as the threshold increases, consistent with the interpretation that noisier observations should exert weaker likelihood guidance.
However, the best guidance strength is not metric-invariant: posterior-standard-deviation error is often minimized at substantially smaller $\lambda$ than posterior-mean error, while MMD, SWD, and spectral error select intermediate regimes.
This separation indicates that a single scalar guidance coefficient cannot simultaneously optimize both mean tendency and uncertainty calibration.

\begin{figure}[t]
  \centering
  \includegraphics[width=\textwidth]{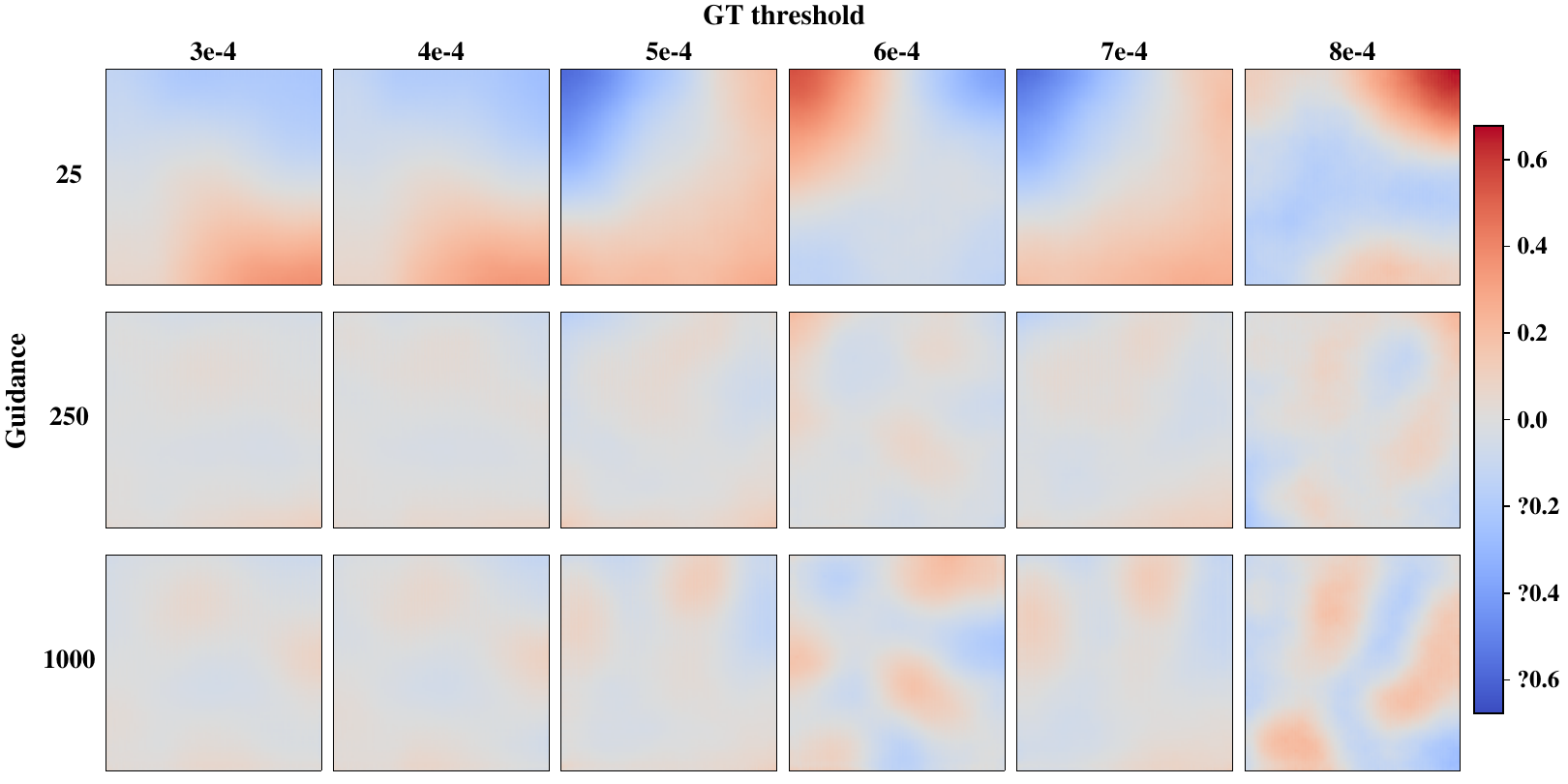}

  \vspace{1.5mm}
  \includegraphics[width=\textwidth]{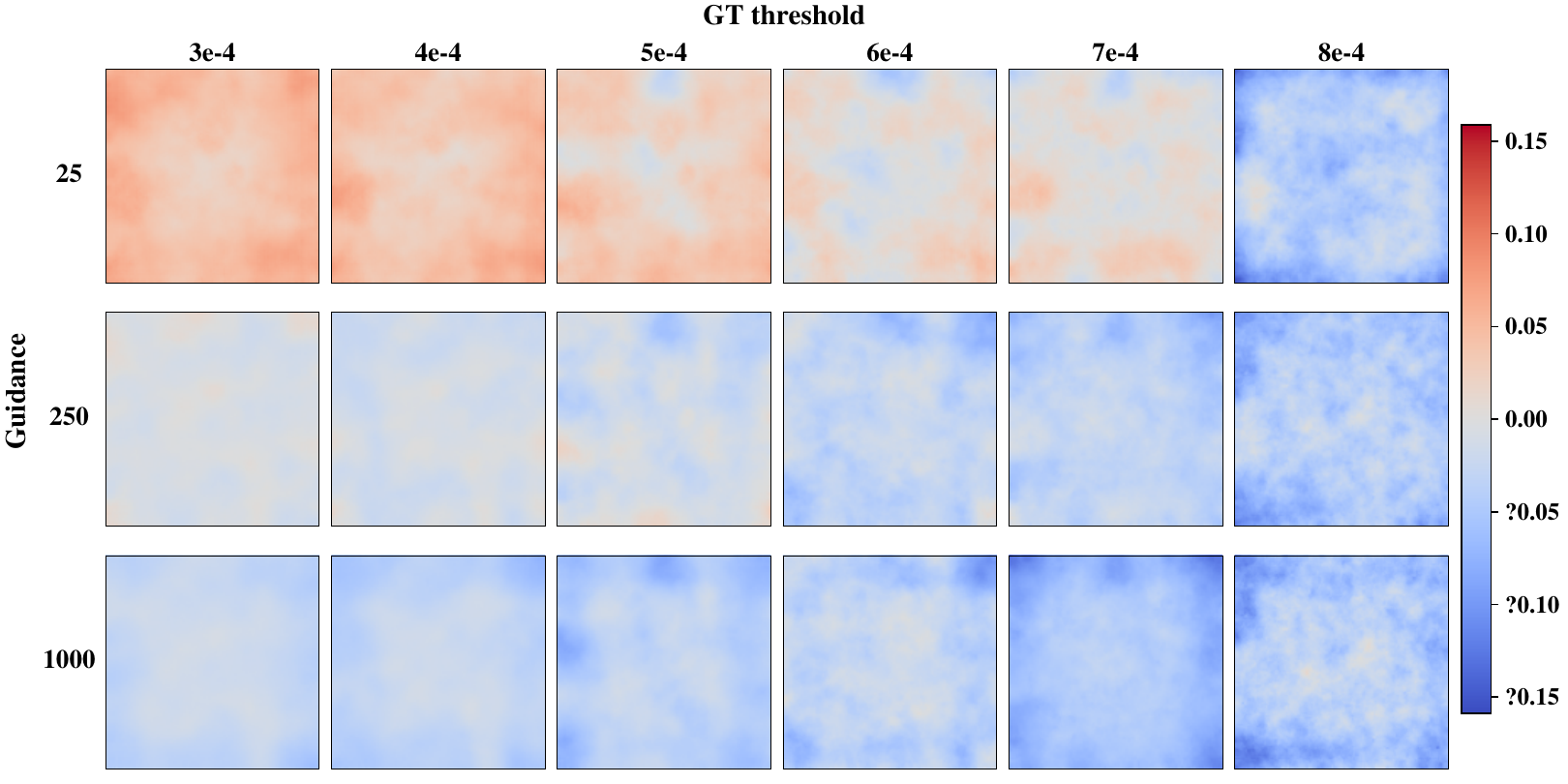}
  \caption{Case-level delta fields for the Poisson guidance sweep.
  The top panel visualizes posterior-mean error fields and the bottom panel visualizes posterior-standard-deviation error fields for the same representative case at representative weak, intermediate, and strong guidance weights.
  These spatial diagnostics illustrate the localized errors induced by under-guidance and over-guidance.}
  \label{fig:poisson_guidance_case_deltas}
\end{figure}

\Cref{fig:poisson_guidance_case_deltas} shows case-level error fields for a representative Poisson source-recovery case at three guidance weights, $\lambda \in \{25,250,1000\}$.
These fields provide a spatial diagnostic of the two main failure modes observed in the sweep.
With weak guidance, the posterior mean retains coherent residual structure because the sampler remains insufficiently conditioned on the observation.
With overly strong guidance, the correction becomes spatially uneven and can introduce localized over-correction, indicating that stronger observation consistency alone does not guarantee a calibrated posterior field.
The standard-deviation panels show the corresponding effect on uncertainty: both under-guidance and over-guidance can distort the spatial distribution of posterior spread.

\clearpage
\subsection{Metric-Pair Diagnostics}
\label{app:metric_pair_diagnostics}

We visualize pairwise relationships among the five main posterior metrics.
Each panel reports raw metric values on log-log axes; consistent trends indicate agreement between two metrics, while scattered panels highlight metric-specific failure modes.

\begin{figure}[htbp]
  \centering
  \includegraphics[width=0.92\textwidth]{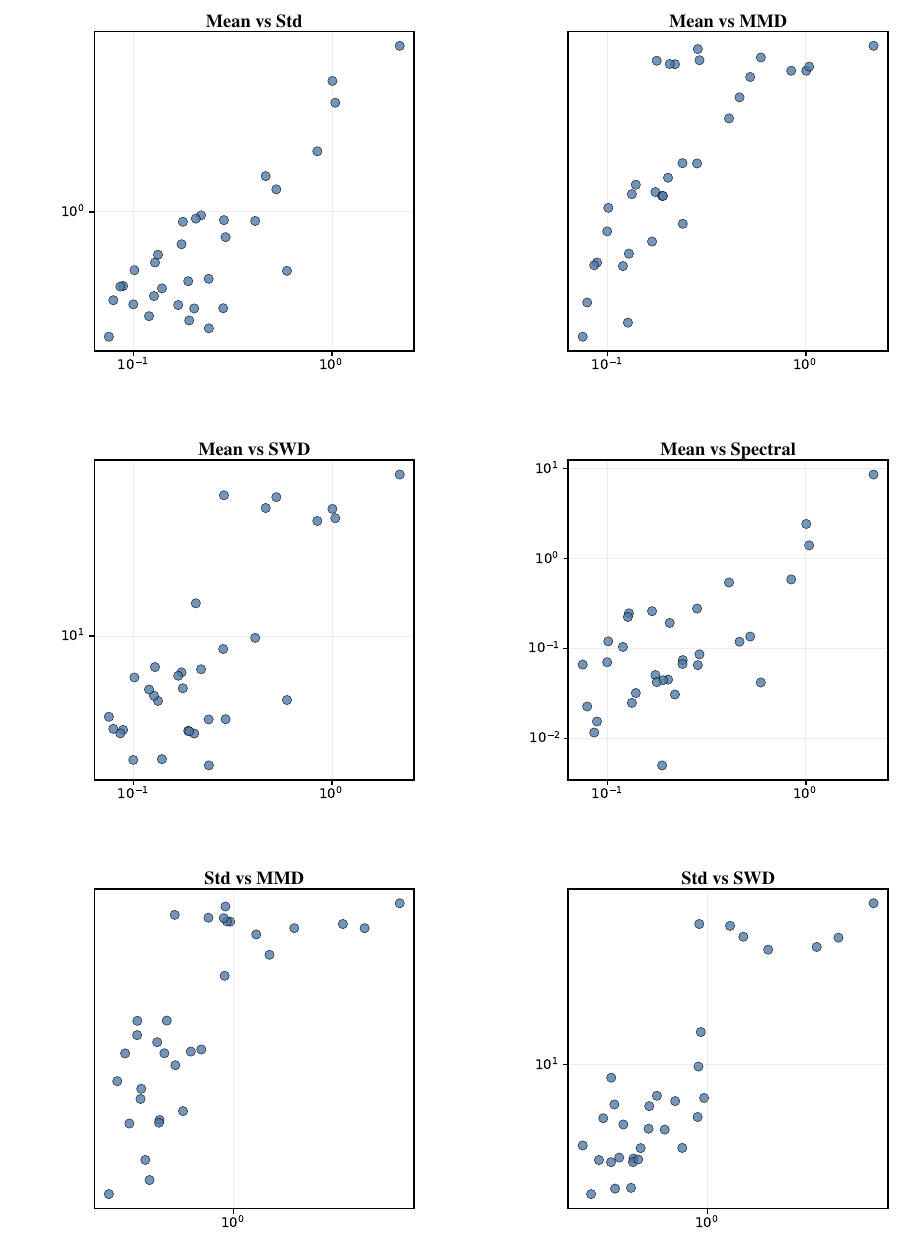}
\end{figure}

\clearpage
\begin{figure}[p]
  \centering
  \includegraphics[width=0.92\textwidth]{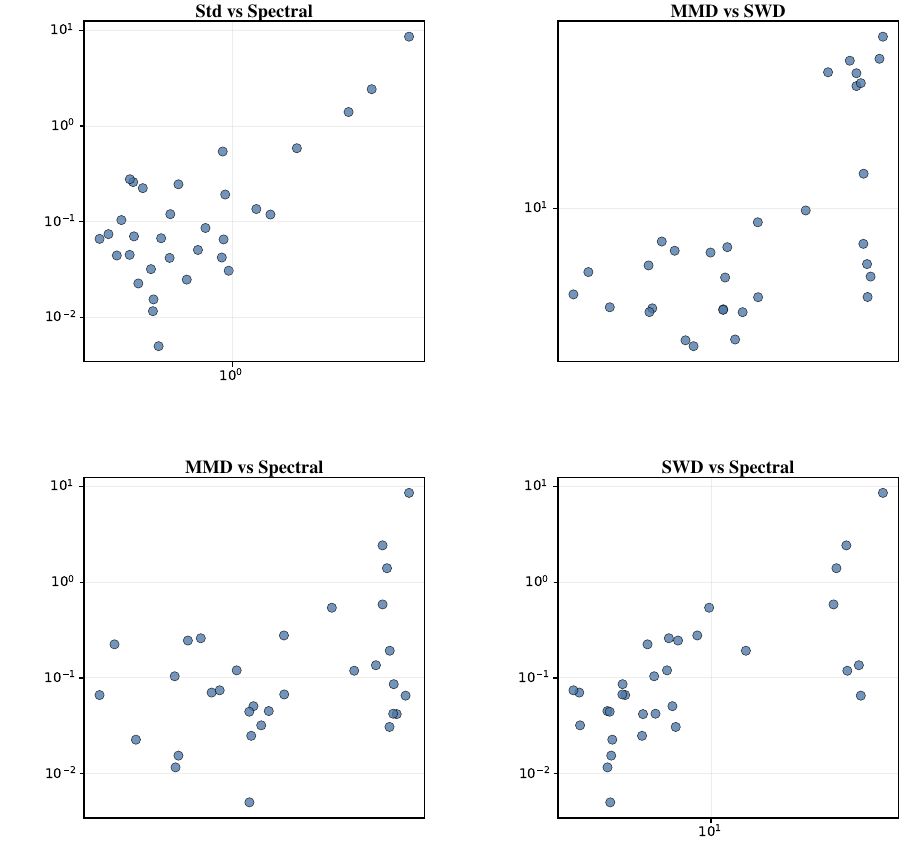}
  \caption{Raw-value metric-pair diagnostics for the five posterior metrics.}
  \label{fig:metric_pair_raw_grid}
\end{figure}

\clearpage
\subsection{Standard Deviations for Main Result Table}
\label{app:variance_tables}

\begin{table}[htbp]
  \centering
  \caption{Standard deviations across cases for the main benchmark metrics in \Cref{tab:main_results}.}
  \label{tab:main_results_std}
  \footnotesize
  \resizebox{\textwidth}{!}{%
  \begin{tabular}{llccccc}
    \toprule
    Task & Method & Mean error & Std error & MMD & SWD & Spectral error \\
    \midrule
    \multirow{7}{*}{\shortstack[l]{Darcy flow\\inversion}} & ECI & 0.1395 & 0.3115 & 0.1027 & 14.2912 & 0.1008 \\
    & ES-MDA & 0.0339 & 0.3445 & 0.0352 & 3.6436 & 0.1899 \\
    & FunDPS & 0.0156 & 0.0605 & 0.0241 & 0.8464 & 0.0820 \\
    & Fun-DDPS & 0.0096 & 0.0660 & 0.0330 & 0.5116 & 0.0653 \\
    & DiffusionPDE & 0.0172 & 0.1412 & 0.0242 & 1.3175 & 0.0822 \\
    & DDIS & 0.0191 & 0.0584 & 0.0283 & 1.5804 & 0.0446 \\
    & FunDiff & 0.0395 & 0.0457 & 0.0474 & 1.6540 & 0.0379 \\
    \midrule
    \multirow{7}{*}{\shortstack[l]{Poisson source\\recovery}} & ECI & 1.4779 & 0.6601 & 0.0118 & 7.2965 & 21.5409 \\
    & ES-MDA & 0.1754 & 0.3814 & 0.0247 & 4.4578 & 5.1581 \\
    & FunDPS & 0.1693 & 0.0640 & 0.0477 & 1.3222 & 0.1573 \\
    & Fun-DDPS & 0.0931 & 0.0469 & 0.0472 & 0.2433 & 0.0567 \\
    & DiffusionPDE & 0.1825 & 0.1459 & 0.0401 & 0.4421 & 0.1710 \\
    & DDIS & 0.0546 & 0.0311 & 0.0265 & 0.3378 & 0.0946 \\
    & FunDiff & 0.1372 & 0.3464 & 0.0319 & 4.9226 & 3.1165 \\
    \midrule
    \multirow{7}{*}{\shortstack[l]{Carbon capture\\and storage}} & ECI & 0.7666 & 2.0413 & 0.0637 & 8.1664 & 257.6435 \\
    & ES-MDA & 0.2960 & 1.8626 & 0.1371 & 4.8350 & 48.8525 \\
    & FunDPS & 0.1093 & 0.2839 & 0.0795 & 2.0215 & 0.9662 \\
    & Fun-DDPS & 0.7437 & 1.7852 & 0.1121 & 4.0355 & 56.4968 \\
    & DiffusionPDE & 0.1608 & 0.2593 & 0.0995 & 3.0374 & 1.3179 \\
    & DDIS & 0.3706 & 0.5733 & 0.1083 & 4.5436 & 7.7058 \\
    & FunDiff & 0.1205 & 0.1492 & 0.0980 & 4.1860 & 0.2191 \\
    \midrule
    \multirow{7}{*}{\shortstack[l]{Light transport\\material inference}} & ECI & 0.0278 & 0.0691 & 0.0362 & 1.4149 & 0.0905 \\
    & ES-MDA & 0.0199 & 0.0235 & 0.0366 & 0.4211 & 0.0302 \\
    & FunDPS & 0.0158 & 0.0238 & 0.0235 & 0.6200 & 0.0216 \\
    & Fun-DDPS & 0.0170 & 0.0201 & 0.0234 & 0.5861 & 0.0891 \\
    & DiffusionPDE & 0.0166 & 0.0201 & 0.0231 & 0.5411 & 0.0873 \\
    & DDIS & 0.0193 & 0.0327 & 0.0334 & 0.6891 & 0.0895 \\
    & FunDiff & 0.0219 & 0.0318 & 0.0291 & 0.7305 & 0.0527 \\
    \bottomrule
  \end{tabular}
  }
\end{table}

\section{Experiment Details}
\label{app:experiment_details}

This appendix records the training and evaluation protocol for the baseline inverse solvers used in PosteriorBench.
All method--dataset pairs are run through the unified repository on NVIDIA B200 GPUs, and measured under the same evaluation pipeline.

\subsection{Baseline Training Protocol}
\label{app:baseline_training}

For learned baselines, the training data for each task consists of paired prior samples and forward observations generated by the task-specific simulator or surrogate described in \Cref{app:data_generation}.
Each method is trained on the official training split for that task and is evaluated only on held-out benchmark cases.
When a method requires a learned prior, score model, flow model, neural operator, or differentiable surrogate, that component is fit without access to the reference posterior samples used for evaluation.
Baseline training and inference configurations are summarized in \Cref{app:baseline_solver_configurations}.

\subsection{Baseline Evaluation Protocol}
\label{app:baseline_evaluation}

At test time, each baseline receives the same observation for a given benchmark case and returns an ensemble of posterior samples.
The reported metrics compare this ensemble with the corresponding reference posterior for that case, using posterior mean error, posterior standard deviation error, maximum mean discrepancy, sliced Wasserstein distance, and radially averaged power-spectrum error.
For fair distributional comparison, methods with larger generated ensembles are subsampled to the common evaluation size used by the task before computing metrics.
All subsampling and metric computation are performed in the physical parameter space of the inverse problem.

\subsection{Baseline Solver Configurations}
\label{app:baseline_solver_configurations}

The remaining subsections summarize the eight posterior solvers used in the benchmark.
For traceability, learned methods are reported with training and, where applicable, inference tables, while ES-MDA is reported by its inference-time assimilation parameters.
The tables focus on solver and architecture parameters; field choices, grid resolutions, training budgets, observation operators, and normalization constants are not treated as method hyperparameters.

\subsubsection{FunDPS}
\label{app:method_config_fundps}

FunDPS~\citep{yao2025guided} trains a joint diffusion prior over the unknown field together with the corresponding observable field.
During posterior sampling, DPS guidance is applied to the observable channel while the joint state is sampled.

\begin{table}[htbp]
  \centering
  \caption{FunDPS training configuration by benchmark task.}
  \label{tab:fundps_training_config}
  \begin{tabular}{llcccc}
    \toprule
    Category & Parameter & Darcy & Poisson & CCS & Light transport \\
    \midrule
    Prior & Architecture & \multicolumn{4}{c}{\texttt{ddpmpp-uno}} \\
     & Learning rate & \multicolumn{4}{c}{$1.0{\times}10^{-4}$} \\
     & LR ramp-up & \multicolumn{4}{c}{5M images} \\
     & EMA half-life & \multicolumn{4}{c}{0.5M images} \\
     & Dropout & \multicolumn{4}{c}{0.13} \\
     & Channel base & \multicolumn{4}{c}{64} \\
     & Channel multipliers & \multicolumn{4}{c}{$[1,2,4,4]$} \\
     & Attention resolutions & $[8]$ & $[16]$ & $[16]$ & $[16]$ \\
     & UNO blocks & \multicolumn{4}{c}{4} \\
     & Operator rank & \multicolumn{4}{c}{0.1} \\
    \bottomrule
  \end{tabular}
\end{table}

\begin{table}[htbp]
  \centering
  \caption{FunDPS inference configuration by benchmark task.}
  \label{tab:fundps_inference_config}
  \begin{tabular}{llcccc}
    \toprule
    Category & Parameter & Darcy & Poisson & CCS & Light transport \\
    \midrule
    Sampling & Initial latent family & \multicolumn{4}{c}{RBF random field} \\
     & Reverse steps & \multicolumn{4}{c}{500} \\
     & $\sigma_{\min}$ & 0.002 & 0.01 & 0.002 & 0.002 \\
     & $\sigma_{\max}$ & 80 & 10 & 80 & 80 \\
     & $\rho$ & \multicolumn{4}{c}{7} \\
    \midrule
    Guidance & Loss & \multicolumn{4}{c}{MSE} \\
     & Field weight & 4000 & 1600 & 300 & 3000 \\
    \bottomrule
  \end{tabular}
\end{table}

\subsubsection{Fun-DDPS}
\label{app:method_config_funddps}

Fun-DDPS~\citep{ju2026funddps} decouples the target-field diffusion prior from the forward map used for likelihood guidance.
The observation map is represented by a task-specific FNO surrogate.

\begin{table}[htbp]
  \centering
  \caption{Fun-DDPS training configuration by benchmark task.}
  \label{tab:funddps_training_config}
  \begin{tabular}{llcccc}
    \toprule
    Category & Parameter & Darcy & Poisson & CCS & Light transport \\
    \midrule
    Prior & Architecture & \multicolumn{4}{c}{\texttt{ddpmpp-uno}} \\
     & Batch size & \multicolumn{4}{c}{32} \\
     & Learning rate & \multicolumn{4}{c}{$1.0{\times}10^{-4}$} \\
     & Attention resolutions & $[8]$ & $[16]$ & $[16]$ & $[16]$ \\
     & Channel base & \multicolumn{4}{c}{64} \\
     & Channel multipliers & \multicolumn{4}{c}{$[1,2,4,4]$} \\
     & UNO blocks & \multicolumn{4}{c}{4} \\
    \midrule
    Surrogate & Architecture & \multicolumn{4}{c}{FNO} \\
     & Fourier modes & $[16,16]$ & $[16,16]$ & $[16,32]$ & $[16,16]$ \\
     & Hidden channels & \multicolumn{4}{c}{48} \\
     & Layers & \multicolumn{4}{c}{4} \\
     & Epochs & \multicolumn{4}{c}{100} \\
     & Batch size & \multicolumn{4}{c}{32} \\
     & Learning rate & \multicolumn{4}{c}{$1.0{\times}10^{-3}$} \\
    \bottomrule
  \end{tabular}
\end{table}

\begin{table}[htbp]
  \centering
  \caption{Fun-DDPS inference configuration by benchmark task.}
  \label{tab:funddps_inference_config}
  \begin{tabular}{llcccc}
    \toprule
    Category & Parameter & Darcy & Poisson & CCS & Light transport \\
    \midrule
    Sampling & Initial latent family & \multicolumn{4}{c}{RBF random field} \\
     & Reverse steps & \multicolumn{4}{c}{500} \\
     & $\sigma_{\min}$ & \multicolumn{4}{c}{0.002} \\
     & $\sigma_{\max}$ & \multicolumn{4}{c}{80} \\
     & $\rho$ & \multicolumn{4}{c}{7} \\
    \midrule
    Guidance & Loss & \multicolumn{4}{c}{MSE} \\
     & Surrogate model & \multicolumn{4}{c}{FNO} \\
     & Field weight & 1000 & 100 & 10.0 & 100.0 \\
    \bottomrule
  \end{tabular}
\end{table}

\subsubsection{DDIS}
\label{app:method_config_ddis}

DDIS~\citep{lin2026decoupled} also separates the target prior from the differentiable forward map, but uses a DAPS-style decoupled sampling update with annealing, diffusion, and Langevin correction phases.
The forward map is represented by a padded FNO surrogate in the canonical configurations.

\begin{table}[htbp]
  \centering
  \caption{DDIS training configuration by benchmark task.}
  \label{tab:ddis_training_config}
  \begin{tabular}{llcccc}
    \toprule
    Category & Parameter & Darcy & Poisson & CCS & Light transport \\
    \midrule
    Prior & Architecture & \multicolumn{4}{c}{\texttt{ddpmpp-uno}} \\
     & Batch size & \multicolumn{4}{c}{32} \\
     & Learning rate & \multicolumn{4}{c}{$1.0{\times}10^{-4}$} \\
     & Attention resolutions & $[8]$ & $[16]$ & $[16]$ & $[16]$ \\
     & Channel base & \multicolumn{4}{c}{64} \\
     & Channel multipliers & \multicolumn{4}{c}{$[1,2,4,4]$} \\
     & UNO blocks & \multicolumn{4}{c}{4} \\
    \midrule
    Surrogate & Architecture & \multicolumn{4}{c}{FNO-pad} \\
     & Fourier modes & \multicolumn{4}{c}{$[64,64]$} \\
     & Hidden channels & \multicolumn{4}{c}{64} \\
     & Layers & \multicolumn{4}{c}{4} \\
     & Epochs & \multicolumn{4}{c}{500} \\
     & Batch size & \multicolumn{4}{c}{40} \\
     & Learning rate & \multicolumn{4}{c}{$1.0{\times}10^{-4}$} \\
    \bottomrule
  \end{tabular}
\end{table}

\begin{table}[htbp]
  \centering
  \caption{DDIS inference configuration by benchmark task.}
  \label{tab:ddis_inference_config}
  \begin{tabular}{llcccc}
    \toprule
    Category & Parameter & Darcy & Poisson & CCS & Light transport \\
    \midrule
    Sampling & Initial latent family & \multicolumn{4}{c}{RBF random field} \\
    \midrule
    Annealing & Steps & 200 & 100 & 100 & 100 \\
     & $\sigma_{\max}$ & \multicolumn{4}{c}{10} \\
     & $\sigma_{\min}$ & \multicolumn{4}{c}{0.01} \\
     & $\rho$ & \multicolumn{4}{c}{7} \\
    \midrule
    Diffusion & Correction steps & \multicolumn{4}{c}{5} \\
     & Correction $\sigma_{\min}$ & \multicolumn{4}{c}{0.001} \\
     & Correction $\rho$ & \multicolumn{4}{c}{7} \\
    \midrule
    Langevin & Steps & 10 & 35 & 20 & 20 \\
     & Learning rate & $1.0{\times}10^{-4}$ & $7.0{\times}10^{-5}$ & $1.0{\times}10^{-4}$ & $1.0{\times}10^{-4}$ \\
     & LR minimum ratio & \multicolumn{4}{c}{0.01} \\
     & LR $\rho$ & \multicolumn{4}{c}{1.0} \\
     & $\eta$ & 0.1 & 0.25 & 0.1 & 0.1 \\
     & $\tau$ & \multicolumn{4}{c}{0.001} \\
    \midrule
    Guidance & Loss & \multicolumn{4}{c}{MSE} \\
     & Surrogate model & \multicolumn{4}{c}{FNO-pad} \\
     & Field weight & 10 & 2.0 & 0.25 & 1.0 \\
    \bottomrule
  \end{tabular}
\end{table}

\subsubsection{DiffusionPDE}
\label{app:method_config_diffusionpde}

DiffusionPDE~\citep{huang2024diffusionpde} trains a joint grid-based score model over the unknown field and the associated physical field.
In our unified runs, observation guidance is applied to the observable channel during reverse diffusion.
We modified the original implementation to fix the normalization issue and support batched inference, which actually improves performance over the stock implementation.

\begin{table}[htbp]
  \centering
  \caption{DiffusionPDE training configuration by benchmark task.}
  \label{tab:diffusionpde_training_config}
  \begin{tabular}{llcccc}
    \toprule
    Category & Parameter & Darcy & Poisson & CCS & Light transport \\
    \midrule
    Prior & Architecture & \multicolumn{4}{c}{\texttt{ddpmpp}} \\
     & Batch size & \multicolumn{4}{c}{32} \\
     & Learning rate & \multicolumn{4}{c}{$1.0{\times}10^{-3}$} \\
     & LR ramp-up & \multicolumn{4}{c}{10M images} \\
     & EMA half-life & \multicolumn{4}{c}{0.05M images} \\
     & Dropout & \multicolumn{4}{c}{0.13} \\
    \bottomrule
  \end{tabular}
\end{table}

\begin{table}[htbp]
  \centering
  \caption{DiffusionPDE inference configuration by benchmark task.}
  \label{tab:diffusionpde_inference_config}
  \begin{tabular}{llcccc}
    \toprule
    Category & Parameter & Darcy & Poisson & CCS & Light transport \\
    \midrule
    Sampling & Initial latent family & \multicolumn{4}{c}{Gaussian} \\
     & Reverse steps & \multicolumn{4}{c}{2000} \\
     & $\sigma_{\min}$ & \multicolumn{4}{c}{0.002} \\
     & $\sigma_{\max}$ & \multicolumn{4}{c}{80} \\
     & $\rho$ & \multicolumn{4}{c}{7} \\
    \midrule
    Guidance & Loss & \multicolumn{4}{c}{L2} \\
     & Observation fraction & \multicolumn{4}{c}{0.8} \\
     & Late-observation multiplier & \multicolumn{4}{c}{0.1} \\
     & Field weight & 40000 & 20000 & 1000 & 400 \\
    \bottomrule
  \end{tabular}
\end{table}

\subsubsection{FunDiff}
\label{app:method_config_fundiff}

FunDiff~\citep{wang2025fundiffdiffusionmodelsfunction} represents target functions with function autoencoders and learns a conditional latent rectified flow with a DiT backbone.
Unlike guidance-based diffusion samplers, FunDiff conditions on the observation representation directly and does not use a scalar likelihood-guidance weight at inference.

\begin{table}[htbp]
  \centering
  \caption{FunDiff training configuration by benchmark task.}
  \label{tab:fundiff_training_config}
  \begin{tabular}{llcccc}
    \toprule
    Category & Parameter & Darcy & Poisson & CCS & Light transport \\
    \midrule
    FAE & Encoder patch size & $16\times16$ & $16\times16$ & $8\times25$ & $8\times8$ \\
     & Embedding dimension & \multicolumn{4}{c}{256} \\
     & Latent tokens & \multicolumn{4}{c}{64} \\
     & Depth & \multicolumn{4}{c}{8} \\
     & Attention heads & \multicolumn{4}{c}{8} \\
     & Max steps & \multicolumn{4}{c}{100k} \\
     & Batch size & \multicolumn{4}{c}{16} \\
     & Peak learning rate & \multicolumn{4}{c}{$1.0{\times}10^{-3}$} \\
     & Target query count & \multicolumn{4}{c}{4096} \\
    \midrule
    DiT & Embedding dimension & \multicolumn{4}{c}{384} \\
     & Depth & \multicolumn{4}{c}{16} \\
     & Attention heads & \multicolumn{4}{c}{8} \\
     & Max steps & \multicolumn{4}{c}{117188} \\
     & Batch size & \multicolumn{4}{c}{128} \\
     & Peak learning rate & \multicolumn{4}{c}{$1.0{\times}10^{-3}$} \\
    \bottomrule
  \end{tabular}
\end{table}

\begin{table}[htbp]
  \centering
  \caption{FunDiff inference configuration by benchmark task.}
  \label{tab:fundiff_inference_config}
  \begin{tabular}{llcccc}
    \toprule
    Category & Parameter & Darcy & Poisson & CCS & Light transport \\
    \midrule
    Sampling & Flow steps & 20 & 20 & 100 & 20 \\
     & Decode chunk & 1024 & 1024 & 2048 & 1024 \\
    \bottomrule
  \end{tabular}
\end{table}

\subsubsection{ECI-sampling}
\label{app:method_config_eci}

ECI-sampling~\citep{cheng2025gradientfree} trains an FNO-based flow-matching model for the joint task state.
At inference, observation consistency is imposed through operator-specific conditioning, including hard replacement for sparse or low-resolution observations where supported by the method profile.

\begin{table}[htbp]
  \centering
  \caption{ECI-sampling training configuration by benchmark task.}
  \label{tab:eci_training_config}
  \begin{tabular}{llcccc}
    \toprule
    Category & Parameter & Darcy & Poisson & CCS & Light transport \\
    \midrule
    Flow model & Backbone & \multicolumn{4}{c}{FNO} \\
     & Fourier modes & \multicolumn{4}{c}{$[32,32]$} \\
     & Hidden channels & \multicolumn{4}{c}{128} \\
     & Layers & \multicolumn{4}{c}{6} \\
     & Embedding channels & \multicolumn{4}{c}{32} \\
     & Base noise & \multicolumn{4}{c}{Matern} \\
    \midrule
    Optimization & Max steps & \multicolumn{4}{c}{58594} \\
     & Batch size & \multicolumn{4}{c}{256} \\
     & Learning rate & \multicolumn{4}{c}{$3.0{\times}10^{-4}$} \\
     & Weight decay & \multicolumn{4}{c}{0} \\
     & Adam $(\beta_1,\beta_2)$ & \multicolumn{4}{c}{$(0.9,0.999)$} \\
    \bottomrule
  \end{tabular}
\end{table}

\begin{table}[htbp]
  \centering
  \caption{ECI-sampling inference configuration by benchmark task.}
  \label{tab:eci_inference_config}
  \begin{tabular}{llcccc}
    \toprule
    Category & Parameter & Darcy & Poisson & CCS & Light transport \\
    \midrule
    Sampling & Steps & \multicolumn{4}{c}{200} \\
     & Mixture count $n_{\mathrm{mix}}$ & 5 & 1 & 1 & 1 \\
     & Resampling interval & 1 & 1 & 5 & 5 \\
    \bottomrule
  \end{tabular}
\end{table}

\subsubsection{ES-MDA}
\label{app:method_config_esmda}

ES-MDA~\citep{Emerick2013b,Jung2018} is a non-neural ensemble smoother baseline.
It does not train a generative model; instead, it updates a task-specific ensemble drawn from the prior using repeated Kalman-style assimilation steps.

\begin{table}[htbp]
  \centering
  \caption{ES-MDA inference configuration by benchmark task.}
  \label{tab:esmda_inference_config}
  \begin{tabular}{llcccc}
    \toprule
    Category & Parameter & Darcy & Poisson & CCS & Light transport \\
    \midrule
    Ensemble & Ensemble size & 128 & 512 & 256 & 1024 \\
    \midrule
    Assimilation & Number of updates & 2 & 1 & 4 & 1 \\
     & Inflation factors $\alpha_k$ & $[2,2]$ & $[1]$ & $[4,4,4,4]$ & $[1]$ \\
    \midrule
    Observation error & Std. model & \multicolumn{4}{c}{prior-predictive standard deviation multiplier} \\
     & Std. multiplier & 10 & 100 & 2 & 1 \\
     & Minimum std & \multicolumn{4}{c}{$1.0{\times}10^{-6}$} \\
     & Jitter & \multicolumn{4}{c}{$1.0{\times}10^{-8}$} \\
    \bottomrule
  \end{tabular}
\end{table}

\subsubsection{FNO with MC Dropout}
\label{app:method_config_mcdropout}

The MC-dropout baseline trains a direct inverse FNO that maps masked observations to the target field.
At inference, dropout layers remain active and posterior samples are obtained from repeated stochastic forward passes of the same trained model.

\begin{table}[htbp]
  \centering
  \caption{FNO with MC Dropout training configuration by benchmark task.}
  \label{tab:mcdropout_training_config}
  \begin{tabular}{llcccc}
    \toprule
    Category & Parameter & Darcy & Poisson & CCS & Light transport \\
    \midrule
    Model & Backbone & \multicolumn{4}{c}{FNO} \\
     & Fourier modes & $[32,32]$ & $[32,32]$ & $[16,64]$ & $[32,32]$ \\
     & Hidden channels & \multicolumn{4}{c}{128} \\
     & Layers & \multicolumn{4}{c}{6} \\
     & Channel-MLP dropout & \multicolumn{4}{c}{0.2} \\
     & Domain padding & \multicolumn{4}{c}{0.1, one-sided} \\
    \midrule
    Optimization & Max images & \multicolumn{4}{c}{15M} \\
     & Batch size & \multicolumn{4}{c}{64} \\
     & Learning rate & \multicolumn{4}{c}{$5.0{\times}10^{-4}$} \\
     & Weight decay & \multicolumn{4}{c}{$2.0{\times}10^{-2}$} \\
     & Scheduler & \multicolumn{4}{c}{One-cycle} \\
    \bottomrule
  \end{tabular}
\end{table}

\end{document}